%% file: main.tex
\documentclass{article} 
\usepackage{iclr2026_conference,times}

\input{math_commands.tex}

\usepackage{hyperref}
\usepackage{url}

\usepackage{graphicx}

\usepackage{subfigure}
\usepackage{mathrsfs}
\usepackage{bm,bbm}
\usepackage{amssymb}
\usepackage{amsmath,amssymb,amsthm}
\usepackage{thmtools}
\usepackage{thm-restate}
\usepackage{tcolorbox}
\usepackage{multirow}
\usepackage{tabularx}
\usepackage{multicol}
\usepackage[commandnameprefix=always]{changes}

\usepackage[capitalize,noabbrev]{cleveref}
\usepackage{amsfonts}
\usepackage{url}
\usepackage{mathtools}
\usepackage{algorithm}
\usepackage{algorithmic}
\usepackage{float}
\usepackage{threeparttable}
\usepackage{graphicx}

\usepackage{wrapfig}

\usepackage{algorithm}
\usepackage{algorithmic}

\newcommand{\fig}{{Figure }}

\newcommand{\eq}{{Eq. }}

\usepackage{algorithmic}
\usepackage{tablefootnote}
\usepackage[dvipsnames]{xcolor}

\newcommand\blfootnote[1]{%
\begingroup
\renewcommand\thefootnote{}\footnote{#1}%
\addtocounter{footnote}{-1}%
\endgroup
}

\usepackage{changes}
\usepackage{balance}
\usepackage{tikz-cd}
\iclrfinalcopy
\title{Memory-Driven Self-Improvement for Decision Making with Large Language Models}

\author{{\bf Xue Yan$^{*~1~2}$ , Zijing Ou$^{~3}$, Mengyue Yang$^{~4}$, 
Yan Song$^{~5}$}\\{\bf Haifeng Zhang$^{^\dag~1~2}$, Yingzhen Li$^{~3}$, Jun Wang$^{\dag~5}$}
\vspace{0.00 cm}\\
{$^1$Institute of Automation, Chinese Academy of Science, Beijing, China}\\
{$^2$School of Artificial Intelligence, University of Chinese Academy of Sciences, China}\\
{$^3$Imperial College London, UK} 
{$^4$University of Bristol, UK}\\
{$^5$AI Centre, Department of Computer Science, University College London, London, UK}}

\begin{document}
\maketitle
\blfootnote{$^*$ Work done during a visit to Imperial College London. $^\dag$Correspondence to: \textlangle haifeng.zhang@ia.ac.cn \textrangle, \textlangle jun.wang@cs.ucl.ac.uk\textrangle}
\vspace{-0.2cm}
\begin{abstract}
Large language models (LLMs) have emerged as effective action policies for sequential decision-making (SDM) tasks due to their extensive prior knowledge.  However, this broad yet general knowledge is often insufficient for specific decision-making tasks with limited task-related data, making it challenging to efficiently adapt LLMs to specific SDM tasks. To address this challenge, we propose a memory-driven self-improvement framework that combines LLM general prior knowledge with a compact memory of domain-specific experiences. Memory retains past interactions and associated Q-values, thereby capturing decision-relevant knowledge that facilitates accurate value estimation and informs the LLM prior refinement. The refined LLM prior, in turn, generates higher-reward trajectories that further enrich memory, forming a natural self-improvement framework where memory and LLM prior mutually reinforce each other. Experiments show that our memory-driven approach significantly outperforms both traditional RL and LLM-based baselines, e.g., improving performance by over 40\% on in-distribution tasks and over 75\% when generalized to unseen tasks in ALFWorld.
\end{abstract}
\section{Introduction}

Sequential decision-making (SDM) has a wide range of real-world applications, including robotics \cite{polydoros2017survey, brunke2022safe, rana2023residual}, autonomous driving \cite{naranjo2005power, song2022quantum}, and human–AI interaction \cite{granter2017alphago, li2019dialogue, mctear2022conversational}. 
Natural language plays a crucial role in many SDM tasks, either in purely language-based settings \cite{granter2017alphago, jannala2025chess, jin2024learning} or as a tool for understanding and describing the environment \cite{ma2024large, wang2025large}. Large language models (LLMs), with their broad prior knowledge, demonstrate strong zero-shot reasoning capabilities, making them promising candidates for action policies in such text-based SDM tasks. However, when deployed in specialized domains, their general knowledge is often insufficient for reliable decision-making \cite{yanefficient, jannala2025chess}. 

To adapt LLM action policies into the target domains, three main approaches have been explored. The first is prompt-based methods, which utilize human-crafted prompts \cite{yao2022react, yao2024tree} or incorporate historical interactions \cite{shinn2024reflexion, christianos2023panguagent} to provide more task-specific information. However, these prompt-engineering methods heavily depend on the quality of the prompts and the reasoning capabilities of the LLMs. The second approach is fine-tuning, which includes supervised fine-tuning (SFT) and reinforcement learning fine-tuning (RLFT). SFT typically requires substantial high-quality decision-making data \cite{zhou2024reflect}, while on-policy RLFT methods \cite{carta2023grounding, tantrue} suffer from poor sample efficiency \cite{abdolmaleki2018maximum, chen2023sufficiency}. The third pipeline performs RL with fixed LLM priors, using LLMs either to narrow the action search space \cite{yanefficient} or to design reward functions \cite{kwonreward, klissarov2023motif} that promote efficient exploration. However, these methods remain highly sensitive to the capability of the LLM priors in fulfilling such roles. 
\begin{figure}[t!]
\vspace{-1mm}
    \centering
\includegraphics[width=1\textwidth]{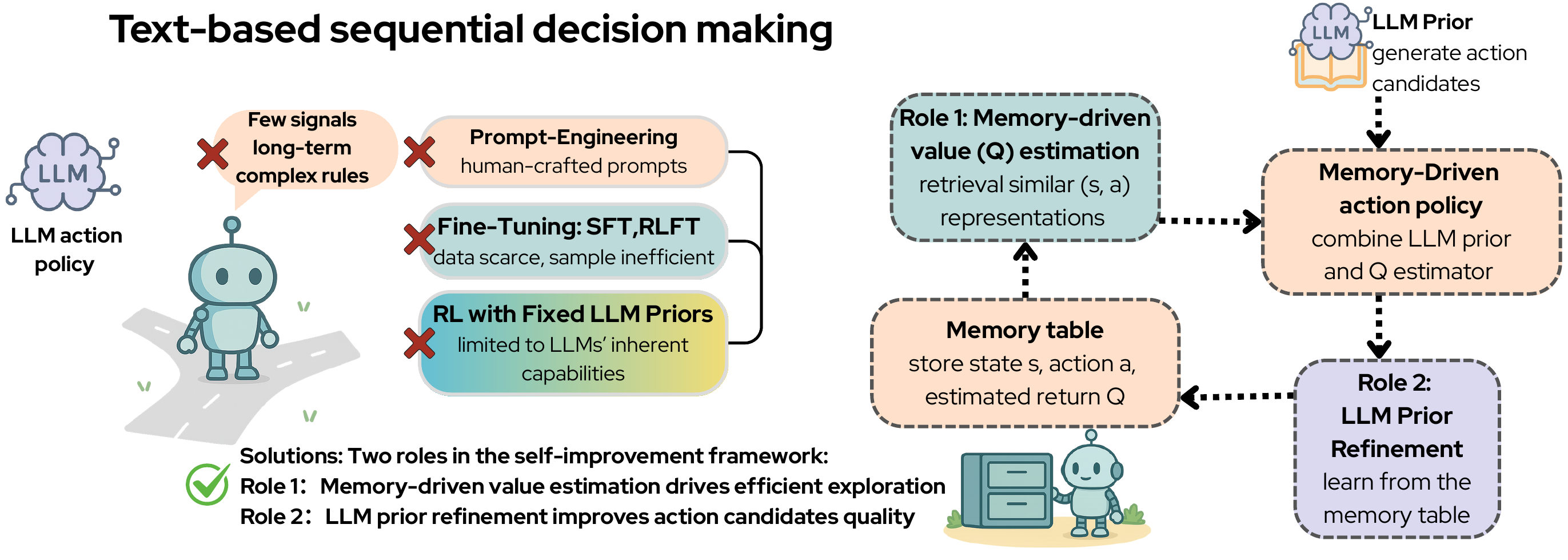}
    \vspace{-2mm}\caption{Motivation and overview of our memory-driven self-improvement framework for text-based SDM. Left: existing approaches (prompt-engineering, fine-tuning, and RL with LLM priors) struggle under sparse signals and domain-specific data. Right: Our framework introduces two complementary roles: (1) memory-driven value estimation, which enables efficient exploration, and (2) LLM prior refinement, which biases action generation toward high-quality candidates; together forming a self-improvement loop that resists scarce experience and enables efficient adaptation.}
    \vspace{-3mm}
    \label{fig:intro}
\end{figure}
\vspace{-0.5mm}

Considering these limitations, we propose a \textbf{memory-driven self-improvement} framework for text-based SDM. To combine the benefits of LLM general knowledge with task-specific interactions, a memory-driven action policy with LLM prior is designed, where the LLM prior generates action candidates, and memory-driven value estimation guides more precise action posterior selection. In practice, the framework forms a closed loop with two mutually reinforcing roles as illustrated in \fig \ref{fig:intro}: \textbf{Role 1}: \textbf{Memory-driven value estimation}, which converts informative interactions into compact memory representations. By retrieving semantically similar past experiences, the model can make non-parametric value estimates for action candidates and enable informed exploration choices. 
\textbf{Role 2}: \textbf{Memory-driven LLM prior refinement}, which periodically updates the LLM's decision prior using historical state-action pairs and their Q-values stored in memory. This refinement biases the LLM prior toward generating high-quality actions, effectively narrowing the search space and improving the convergence rate. 
Overall, the informative memory table provides a reliable foundation for LLM prior refinement, while the refined LLM prior leads to higher-quality actions that further enrich the memory, thus naturally constructing a self-improvement framework. This mutual reinforcement enables scalable and efficient adaptation to target SDM tasks.

In summary, our main contributions are as follows:

1. We propose a \textbf{memory-driven self-improvement framework} for text-based SDMs and leverage the Expectation-Maximization (EM) to provide a unified formulation and practical implementation.

2. We introduce a \textbf{memory-driven value estimation} approach that utilizes LLMs’ representation capabilities and retrieval techniques to achieve meaningful, non-parametric Q-value estimation.

3. We present a \textbf{memory-driven policy optimization} method that defines a powerful action policy as the combination of LLM priors and memory-based Q-estimation, and uses experiences stored in memory to refine the LLM prior. 

4. Experimental results on ALFWorld and Overcooked demonstrate that memory-driven value estimation achieves superior sample efficiency, while LLM prior refinement proves crucial for further expanding the capabilities of LLM-based action policies.

\section{Preliminary}
\label{pre}
\textbf{Textual Markov Decision Processes.}
A Markov Decision Processes (MDP) is defined as $\langle\mathcal{S}, \mathcal{A}, \mathcal{P}, \mathcal{R}, \gamma\rangle$, where $\mathcal{S}$ is the state space, $\mathcal{A}$ is the action space, $\mathcal{P}: \mathcal{S}\times\mathcal{A}\rightarrow\mathcal{S}$ is the transition function, $\mathcal{R}:\mathcal{S}\times\mathcal{A}\rightarrow\mathbb{R}$ is the reward function, and $\gamma\in(0,1)$ is the discount factor. Particularly, note that $r_t$ denotes the reward received at time step $t$. 
In this work, we focus on textual MDP, where both states and actions are represented in natural language, i.e., $\mathcal{S},\mathcal{A}\subseteq\mathcal{V}^*$, with $\mathcal{V}$ denoting the vocabulary. 
Textual MDPs present unique challenges, as the state and action spaces can be combinatorially large, and policies must operate over inherently discrete, structured, and semantically rich representations of language. 

\textbf{Q-learning for MDPs} 
A prominent paradigm for solving MDPs with discrete action spaces is Q-learning, where the Q-function $Q(s,a)$ is learned to rank and select actions. For each state–action pair $(s,a)$, the Q-function, defined as $Q(s,a) = \mathbb{E}_\pi\left[\sum_{i\geq t} \gamma^{i-t} r_i \mid s, a\right]$, represents the expected cumulative reward obtained by starting from $(s,a)$ and following policy $\pi$ thereafter. The DQN algorithm \citep{mnih2013playing} is a classic instance of Q-learning that maps state–action embeddings to scalar Q-values and trains the Q-network via temporal-difference (TD) learning. \cite{blundell2016model} introduces Episodic Control(EC), a memory-based Q-learning method that leverages retrieval techniques and informative state representations to enable non-parametric Q-value estimation. 

\textbf{Control as Inference.}
Control-as-Inference framework \citep{levine2018reinforcement} formulates the solution of MDPs from a probabilistic inference perspective. An optimality random variable $\mathcal{O}$ is introduced, where $\mathcal{O}=1$ indicates achieving a successful trajectory and $\mathcal{O}=0$ otherwise. 
The objective is to maximize the likelihood of achieving a successful trajectory from each state $s$, which is identical to maximizing the evidence lower bound (ELBO):
\begin{equation}
\label{elbo}
\log p(\mathcal{O} \!=\! 1 | s) \geq \mathbb{E}_{q(a|s)}[\log p(\mathcal{O} \!=\! 1|s,a)] \!-\! D_\text{KL}(q(a|s)\|p(a|s)) \!\triangleq\! \mathrm{ELBO},
\end{equation}
where $p(a|s)$ is the prior action distribution, $q(a|s)$ is the variational distribution, and $p(\mathcal{O}=1|s,a)$ is the likelihood that the trajectory will achieve optimality given the current state-action pair $(s,a)$. \cite{abdolmaleki2018maximum} assume that the optimality likelihood is proportional to Q-value: $p(\mathcal{O}=1|s,a) \propto \exp(Q(s,a)/\alpha)$. 
This probabilistic formulation offers both conceptual flexibility and methodological richness, while also naturally accommodating the integration of LLMs as sources of prior knowledge \citep{yanefficient}. 


\section{Memory-Driven value estimation with LLM priors}
\label{likelihood}
In this section, we present a memory-driven approach to Q-value estimation that leverages LLM-based semantic representations. By explicitly exploiting LLM representations through retrieval techniques, our method builds upon the well-established episodic control (EC) \citep{blundell2016model} to enable non-parametric Q-value estimation. Specifically, our method maintains a memory table $\mathcal{M}$ storing Q-values for visited state–action pairs, which is continuously updated during online exploration. At inference time, actions are selected through memory queries. The framework is thus defined by two core operations: \textbf{memory update} and \textbf{memory query}.

\begin{wrapfigure}{r}{0.55\linewidth}
\vspace{-6mm}
\centering
\begin{minipage}{0.99\linewidth}
    \begin{algorithm}[H]
    \caption{Memory-Driven Q-learning (Mem-Q)} \small
    \label{alg:train-proposal} 
    \begin{algorithmic}[1] 
        \REQUIRE Embedding function $f_\text{LLM}$, empty memory table $\mathcal{M} = \emptyset$. 
    \ENSURE Updated memory table $\mathcal{M}$.
    \FOR{each episode}
        \FOR{$t = 1,2,3,\ldots,T$}
            \STATE \textcolor{orange}{Obtain all feasible actions  $\mathcal{A}_{s_t}$ for current $s_t$.t}
            \STATE Estimate the return $\widehat{Q}$ for each $a \in \mathcal{A}_{s_t}$ via~\eq \ref{mem_Q}.
            \STATE \textcolor{orange}{Select action $a_t$ using the $\epsilon$-greedy strategy.} 
            \STATE Execute action $a_t$, receive reward $r_{t+1}$, and observe the next state $s_{t+1}$.
        \ENDFOR
        \FOR{$t = T, T-1, \ldots, 1$}
            \STATE Write and update memory table $\mathcal{M}$ via \eq \ref{qupdate}.
        \ENDFOR
    \ENDFOR
    \end{algorithmic}
    \end{algorithm}
\end{minipage}
\vspace{-5mm}
\end{wrapfigure}
\textbf{Memory update.} A memory table $\mathcal{M}$ is constructed to store information about previously visited state-action pairs $(s, a)$. Specifically, it includes the natural language descriptions of $(s, a)$, the corresponding vectorized embeddings $f(s,a)$, and the associated Q-values $Q(s,a)$. The Q-values stored in memory $\mathcal{M}$ are then updated according to:
\begin{equation} \label{qupdate}
\scalebox{0.85}{
$\!\!\!\!\!\!\!\!Q(s_t, a_t) \!\!\leftarrow\!\! \begin{cases}
R_t,  \hfill \text{if } (s_t, a_t) \notin \mathcal{M},  \\
\max\{Q(s_t, a_t), R_t\},  \text{otherwise},\!\!\!\!\!\!\!\!\!\!
\end{cases}$
}
\end{equation}
where $R_t = \sum_{i\geq t} \gamma^{i-t} r_i$ represents a Monte Carlo estimate of the cumulative discounted reward (i.e., the return-to-go).
 
\textbf{Memory query.} Given the memory $\mathcal{M}$ and the current state $s_t$, the policy is then determined using  the kernel-based Q-value estimator $\widehat{Q}$, which is defined as
\begin{align} \label{mem_Q}
    \widehat{Q}(s_t, a) \!\!\!=\!\!\!\! \sum_{i \in \mathcal{N}_{M} (f(s_t, a))}\!\!\!\!\!\!\! w_i Q(s^{(i)}, a^{(i)}), \ \ w_i = \frac{k(h, h_i)}{\sum_{j\in \mathcal{N}_{M}} k(h, h_j)}, h = f(s_t, a), h_i = f(s^{(i)}, a^{(i)}),
\end{align}
where $\mathcal{N}_{M}(f(s_t, a))$ denotes the $M$ nearest neighbors of $f(s_t, a)$, and the inverse distance kernel is used with $k(h, h_j) = \frac{1}{\|h - h_j\|_2^2 + \delta}$, which measures similarity in the embedding space.
With the estimated Q-values $\widehat{Q}$, action selection can be carried out using the $\epsilon$-greedy strategy, analogous to the approach in DQN \citep{mnih2013playing}. 



\begin{wrapfigure}{r}{0.53\linewidth}
  \centering
  \vspace{-2mm}
  \begin{minipage}{0.49\linewidth}
      \includegraphics[width=0.99\linewidth]{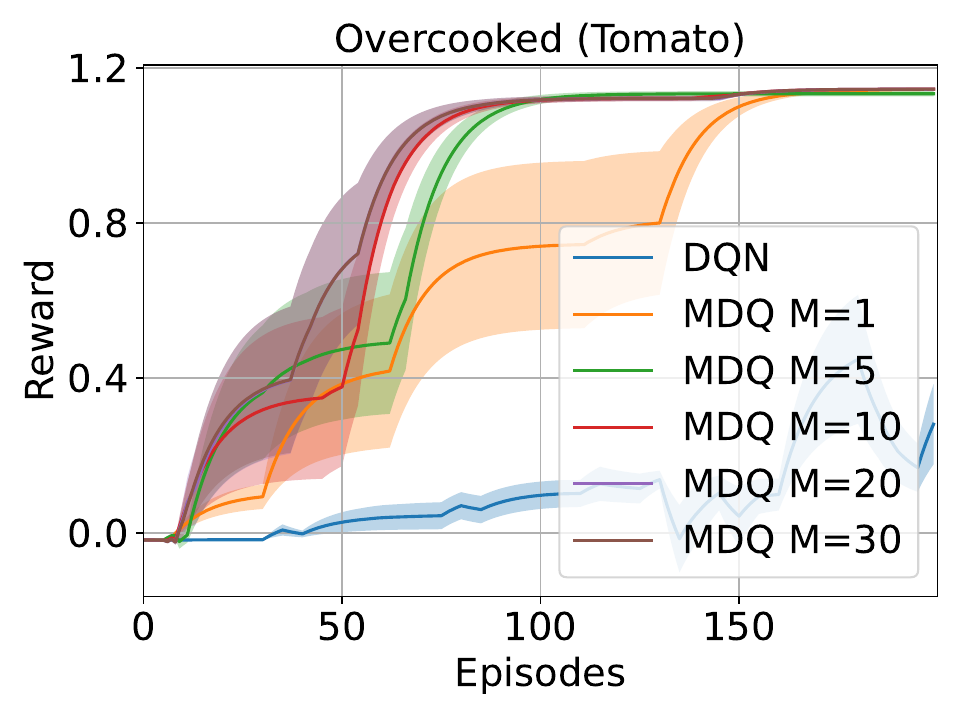}
  \end{minipage}
  \begin{minipage}{0.49\linewidth}
      \includegraphics[width=0.99\linewidth]{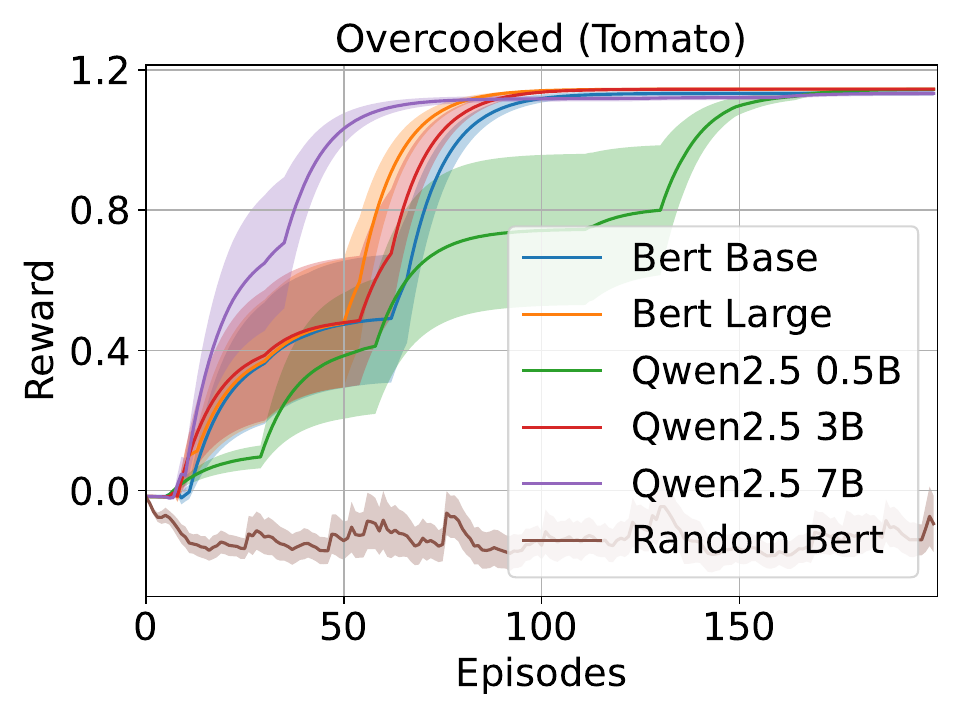}
  \end{minipage}
  \vspace{-4mm}
  \caption{Results of memory-drive Q-learning on Overcooked. Left: effect of the number of retrieved $(s,a)$ pairs for value estimation; Right: effect of different LLMs on representations.}  
    \label{fig:ec-overcook-llm-prior}
  \vspace{-2mm}
\end{wrapfigure}
\paragraph{Vectorized Memory Query with LLM Priors.}
While conceptually simple, the effectiveness of the proposed memory-driven Q-learning critically depends on the choice of embedding function $f$.
Inspired by the recent success of Retrieval Augmented Generation (RAG) \citep{wiratunga2024cbr}, which demonstrates the ability of LLMs to yield semantically rich vectorized representations, we leverage LLM-based embeddings for $(s,a)$ pairs to enhance retrieval quality and improve value estimation. Specifically, we adopt the encoder of the LLM to be the embedding function $f_\text{LLM}$. The overall memory-driven decision-making procedure with LLM embedding function is summarized in \cref{alg:train-proposal}, referred as Memory-Driven Q-learning(Mem-Q). 

In \cref{fig:ec-overcook-llm-prior}, we evaluate the Mem-Q in Overcooked  \citep{tantrue}, a textual decision-making task. It shows that our Mem-Q significantly outperforms DQN. The ablation study on the retrieval size $M$, introduced in \eq \ref{mem_Q}, shows that incorporating more similar $(s,a)$ embeddings in the kernel-based Q estimation leads to faster convergence, which may be attributed to the more accurate value estimation using more meaningful representations \cite{han2023explaining}. Furthermore, larger LLMs, which capture richer semantic structures, consistently yield stronger performance.  

\section{Memory-Driven policy optimization with LLM priors} \label{action}

The success of memory-driven Q-learning highlights a key insight: combining the capabilities of LLMs with experience storage can substantially enhance the efficiency of RL algorithms. Motivated by this, we propose a \textbf{memory-driven action policy} that uses the LLM prior to narrow the action search space, and leverage memory to further refine the LLM prior with domain-specific knowledge, thereby improving sample efficiency.


\subsection{Memory-Driven Policy with LLM Priors}
\label{prior policy}
A straightforward approach to incorporating LLMs into the action policy is through a probabilistic inference framework \citep{li2024q}, wherein the policy is cast as the posterior distribution:
\begin{align}
    p_\theta(a|s,\mathcal{O}=1) 
    \propto p(\mathcal{O}=1|s,a)p_{\text{LLM}_\theta}(a|s),
\end{align}
where $p_{\text{LLM}_\theta}(a|s)$ is the LLM prior parametrized by $\theta$ and $p(\mathcal{O}=1|s,a)$ denotes the likelihood of optimality\footnote{Notably, the likelihood exhibits no explicit parametric dependence, as the proposed memory-driven Q-learning framework is inherently non-parametric.}, which can be estimated using Q-value, as described in Eq~\ref{mem_Q}.
Therefore, the memory-driven policy can be conducted using self-normalized importance sampling as follows:
\begin{itemize}
    \item Sample $K$ candidate actions from the LLM prior: $\mathcal{C}^K(s)\!=\!\{a_1,\ldots,a_K\}, a_k \!\sim\! p_{\text{LLM}_\theta}(\cdot|s)$.
    \item For each action candidate $a_k$, approximate the optimality likelihood using the kernal-based Q-value estimator $\widehat{Q}(s,a_i)$ according to Eq~\ref{mem_Q}.
    \item Select an action via: $a \sim \mathrm{Multinormial}\left( \exp(\widehat{Q}(s,a_1)/\tau) / \sum_k \exp(\widehat{Q}(s,a_k) / \tau) \right)$.
\end{itemize}
Although conceptually straightforward, this method is sensitive to the specification of the LLM prior $p_\theta$. A poorly aligned prior may bias the candidate set toward suboptimal actions, thereby degrading policy performance. To overcome this limitation, the following section introduces a principled optimization procedure based on the Expectation–Maximization framework, which refines the memory-driven policy by adaptively optimizing the underlying LLM priors. 

\begin{algorithm}[t!]
\begin{threeparttable}

\caption{Memory-Driven Expectation Maximization (Mem-EM) \textcolor{ForestGreen}{with LLM Prior Refinement}}
\label{alg:Mem-EM}
\begin{algorithmic}[1]
    \REQUIRE LLM action prior $p_{\text{LLM}_{\theta}}$, memory table $\mathcal{M}=\emptyset$
    \ENSURE Refined LLM prior $p_{\text{LLM}_{\theta}}$ and memory table $\mathcal{M}$
    \FOR{episode $i = 1$ to $N$}
        \FOR{step $t = 1,2,\ldots,T$}
            \STATE \textcolor{orange}{Sample action candidates from LLM prior $\mathcal{C}^K(s_t) \sim p_{\text{LLM}_\theta}(\cdot|s_t)$.}
            \STATE Estimate Q-values $\widehat{Q}(s_t,a)$ for $a \in \mathcal{C}^K(s_t)$ via~\eq \ref{mem_Q}.
            \STATE \textcolor{orange}{Select action $a \sim \mathrm{Multinormial}( \exp(\widehat{Q}(s,a_1)/\tau) / \sum_k \exp(\widehat{Q}(s,a_k) / \tau) )$}.
            \STATE Execute $a_t$, observe reward $r_{t+1}$ and next state $s_{t+1}$
        \ENDFOR
        \FOR{step $t = T$ down to $1$}
            \STATE Write and update memory table $\mathcal{M}$ via \eq \ref{qupdate}. 
        \ENDFOR
        \IF{{$i$ reaches update interval}}
    \STATE \textcolor{ForestGreen}{Refine LLM prior $p_{\text{LLM}_{\theta}}$ using stored $(s,a)$ pairs from $\mathcal{M}$ via Eq.~\ref{priorlearn}.}
        \ENDIF
    \ENDFOR
\end{algorithmic}
\end{threeparttable}
\vspace{0.3em}
\noindent\rule{\textwidth}{0.4pt}
\vspace{0.3em}
\begin{minipage}{\textwidth}
\footnotesize
\textbf{Note:} \textcolor{orange}{Orange text} highlights LLM-guided exploration steps that differ from the Mem-Q in \cref{alg:train-proposal}; \textcolor{ForestGreen}{green text} indicates prior refinement operations.
\end{minipage}
\vspace{-0.5em}
\end{algorithm}
\vspace{-1em}

\subsection{Optimizing Memory-Driven Policy with Expectation Maximization} 
To optimize the memory-driven policy with the LLM prior and the likelihood estimation involved, we adopt the Expectation–Maximization (EM) algorithm. 
Starting from an arbitrary initialization $\theta_0$, the EM procedure performs the following iterative update:
\begin{align} \label{eq:em-update}
    \theta_{k+1} = \argmax_{\theta} \E_{p_{\theta_k}(a,s | \mathcal{O}=1)} \left[ \log p(\mathcal{O}=1|s,a) + \log p_{\text{LLM}_\theta} (a|s) \right].
\end{align}
By construction, the EM update can be interpreted as maximizing the ELBO in Eq.~\ref{elbo}.
In practice, however, the expectation in Eq.~\ref{eq:em-update} is analytically intractable. To address this challenge, we employ an on-policy E-step for likelihood expectation approximation and a memory-driven off-policy E-step for prior expectation approximation, followed by maximization based on these tractable estimates. 

\textbf{Expectation Step.} 
In the E-step, a simple yet tractable approach for the expectation approximation is the Monte Carlo (MC) estimation. Concretely, for the current state $s$, we draw an action from the posterior $a \sim p_{\theta} (a | s, \mathcal{O}=1)$, which can be approximated using importance sampling introduced in Sec. \ref{prior policy}. For the likelihood expectation term in \eq \ref{eq:em-update}, the likelihood is assumed to be proportional to the Q value, and one can directly apply the kernel-based Q value estimation in \eq \ref{mem_Q}.

However, for the LLM prior expectation term, such an ``on-policy'' approach is inefficient, since LLMs typically involve a very large number of parameters, and a limited set of MC samples is insufficient to provide reliable gradient estimates for subsequent M-step.  Consequently, we consider importance sampling approximation using examples stored in the memory table to explore and exploit the posterior distribution effectively, thereby yielding a robust ``off-policy" estimate of the expectation. 
Specifically, we employ self-normalized importance sampling (SNIS) to estimate the LLM prior expectation as follows:
\begin{align}
    \E_{p_\theta (a, s | \mathcal{O}=1)}\left[ \log p_{\text{LLM}_\theta}(a|s) \right] 
    &= \E_{q(s,a)} \left[ \frac{p_\theta (a, s | \mathcal{O}=1)}{q(s,a)} \log p_{\text{LLM}_\theta}(a|s) \right] \nonumber \\
    &\approx \sum_i \frac{w(s^{(i)}, a^{(i)})}{\sum_j w(s^{(j)}, a^{(j)})} \log p_{\text{LLM}_\theta}(a^{(j)}|s^{(j)}), \ (s, a) \sim q(s,a) \nonumber
\end{align}
where $w(s,a) = \frac{p(\mathcal{O}=1|a,s) p_{\text{LLM}_\theta}(a,s)}{q(s,a)}$ denotes the importance weight.
While the proposal $q(s,a)$ offers substantial flexibility, the statistical efficiency of SNIS is highly sensitive to its choice: suboptimal proposals yield high-variance importance weights, thereby impeding effective state–action exploration. In theory, the optimal proposal, which leads to zero variance, is $q(s,a) \propto p(\mathcal{O}=1|a,s) p_{\text{LLM}_\theta}(a,s)$. 
While intractable, empirically, the memory table $\mathcal{M}$ provides a practical basis for designing the proposal distribution $q(s, a)$, as it stores state-action pairs accumulated from previous posterior samples. We thus adopt a uniform distribution over the memory table $\mathcal{M}$ as the proposal, providing an empirical approximation to the posterior. Specifically, the importance weight can be approximated as:
\begin{align}
    w(s,a) = \frac{p(\mathcal{O}=1|a,s) p_{\text{LLM}_\theta}(a,s)}{q(s,a)} 
    \approx \frac{p(\mathcal{O}=1|a,s) q(a,s)}{q(s,a)}
    \propto \exp(Q(s,a) / \tau)
\end{align} 
where we further use the proposal $q(s,a)$, which is the emperical distribution of the memory table $\mathcal{M}$, to approximate the joint LLM prior $p_{\text{LLM}_\theta}(s,a)$. 
Although heuristic, this approximation is empirically found to stabilize training. Intuitively, at iteration $k$, the optimal LLM prior coincides with the posterior from the previous step, $p_{\theta_k}(a|s,\mathcal{O}=1)$, as suggested by Eq.~\ref{eq:em-update}.
Hence, the intractable LLM prior can be approximated by the empirical distribution $q(s,a)$, functioning as a moving average of the posterior.
This approach eliminates the need for repeated LLM queries to estimate the prior, thereby reducing computational cost and improving training efficiency. 


\textbf{Maximization Step.}
In the M-step, we maximize the expectation in Eq.~\ref{eq:em-update}, which involves the optimality likelihood $p(\mathcal{O}=1 \mid s,a)$ and the LLM prior $p_{\text{LLM}_\theta}$.
To maximize the expectation of the non-parametric likelihood, we update the memory according to Eq.~\ref{qupdate} using the sample drawn from the posterior. For optimizing the LLM prior, building on the memory-driven expectation estimation, we apply the following memory-based reweighted training objective:
\begin{align}
\label{priorlearn}
    \theta_{k+1} = \argmin_\theta \sum_{(s,a) \sim \mathcal{M}} \left[ \frac{\exp(Q(s,a) / \tau)}{\sum_{(s^\prime,a^\prime) \sim \mathcal{M}} \exp(Q(s^\prime,a^\prime) / \tau)} \log p_{\text{LLM}_\theta} (a|s) \right].
\end{align}
We summarize the detailed training procedure in Algorithm~\ref{alg:Mem-EM}. It is noteworthy that memory-driven Q estimation and memory-driven policy optimization interact in a bootstrapping manner. Specifically, the memory-driven Q estimation module continuously refines the non-parametric estimate of the optimality likelihood, which in turn provides more accurate feedback for updating the LLM prior. Conversely, the improved LLM prior constrains the action search space toward high-quality candidates, thereby accelerating convergence and improving the sample efficiency of memory-driven Q estimation. This closed-loop interaction establishes a self-improvement cycle, where the two components iteratively enhance one another, leading to progressively stronger policies.

\section{Experiments}
In this section, we design experiments to validate the effectiveness of our memory-driven self-improvement framework.
\label{sec:exp}
\subsection{Environments}
We consider two textual decision-making tasks:
\newline\textbf{Overcooked} The textual Overcooked environment \citep{tantrue} benchmarks the ability of taking a sequence of actions to deliver dishes. We consider two Overcooked tasks: Overcooked(Tomato), which requires delivering a dish of chopped tomato, and Overcooked(Salad), which requires delivering a salad containing chopped tomato and lettuce. The feasible actions vary with state changes, and the maximum number of feasible actions for one state is 8. Besides the reward of 1 for successfully delivering a dish, this environment also provides the following reward shaping: +0.2 for correctly chopping an ingredient, +1 terminal reward for successfully delivering the correct dish, -0.1 for delivering any incorrect item, and -0.001 for each time step.
\newline\textbf{ALFWorld} \citep{shridhar2020alfworld} is a popular and complex decision-making benchmark for navigating and completing tasks within rooms. Both observations and actions are textual, and the action space consists of high-level plans such as "go to a room" that can be understood using LLMs' prior knowledge. The action space is finite but large, with varying feasible action sets for each state, and the maximum possible admissible action space for one step reaches up to 50, making exploration from scratch challenging. This benchmark contains thousands of subtasks, making it convenient for testing generalization performance on unseen tasks. We consider two classes of ALFWorld tasks: ALFWorld(pick) and ALFWorld(examine). There are no auxiliary rewards except for a reward of 1.0 for reaching the final goal.
\begin{figure*}[t!]
\vspace{-2mm}
\centering
	  \includegraphics[width=0.95 \linewidth]{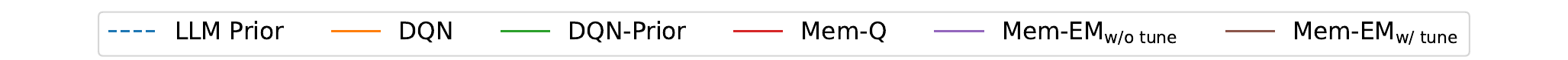}\\
{\includegraphics[width=0.325\linewidth ]{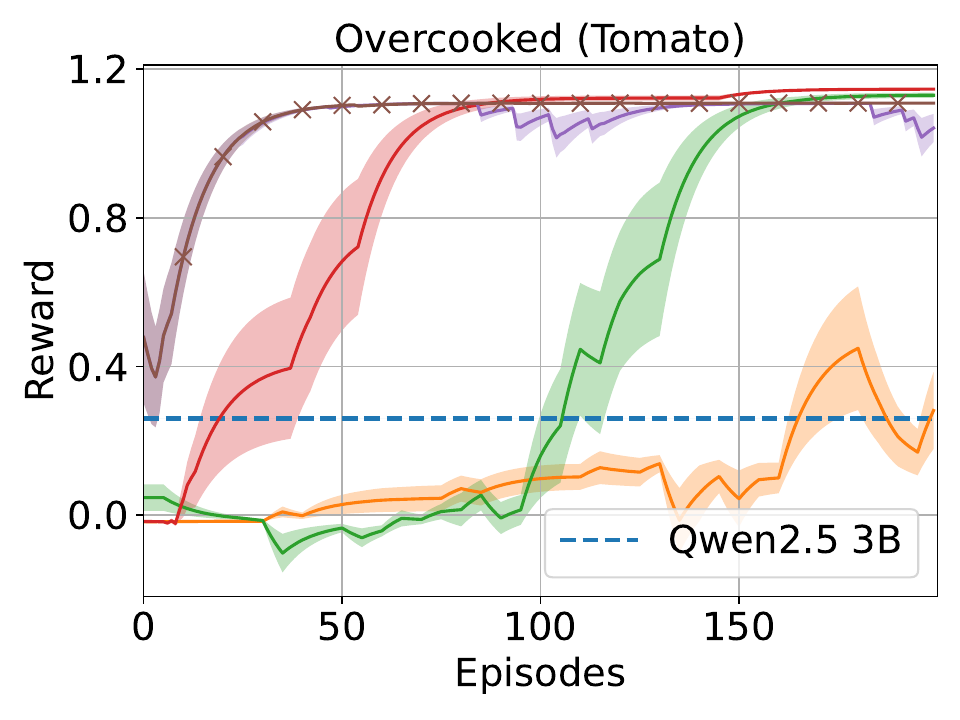}}
{\includegraphics[width=0.325\linewidth ]{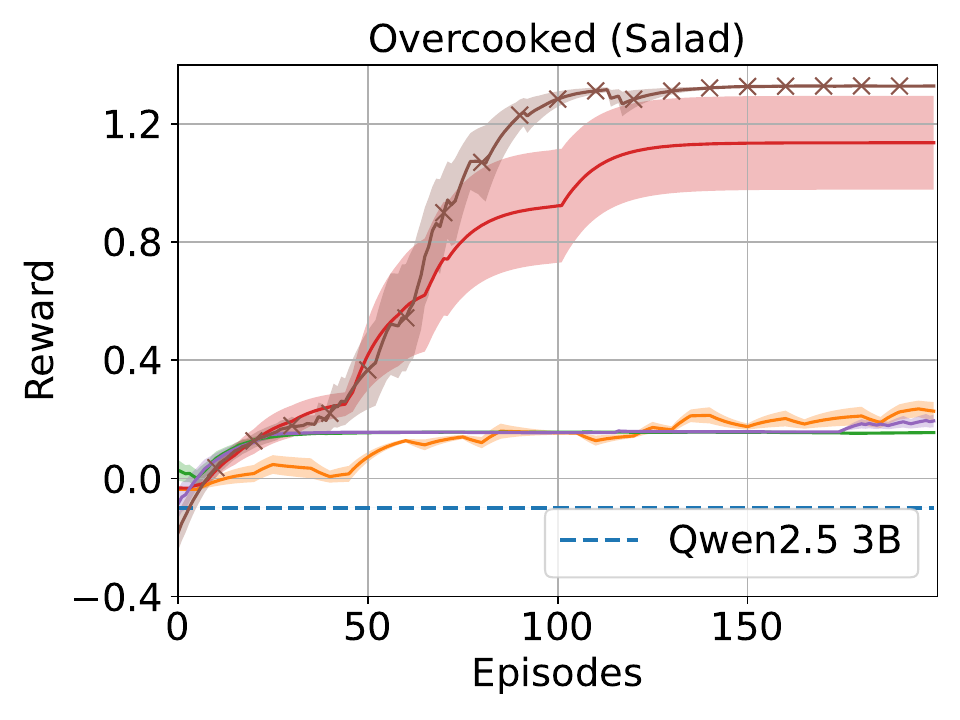}}
{\includegraphics[width=0.325\linewidth ]{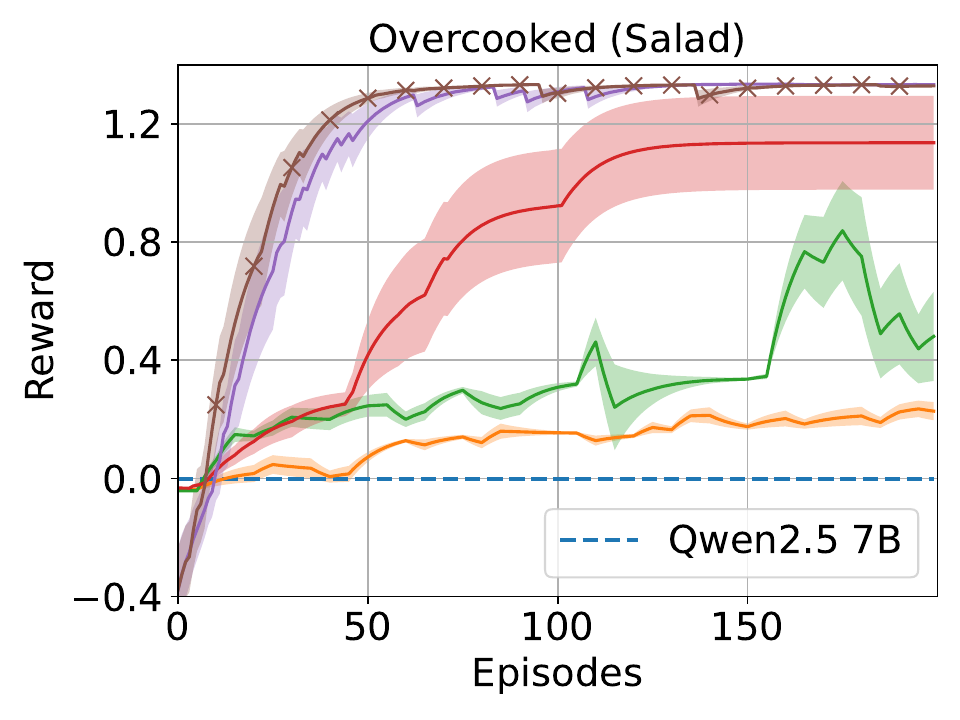}}
{\includegraphics[width=0.325\linewidth ]{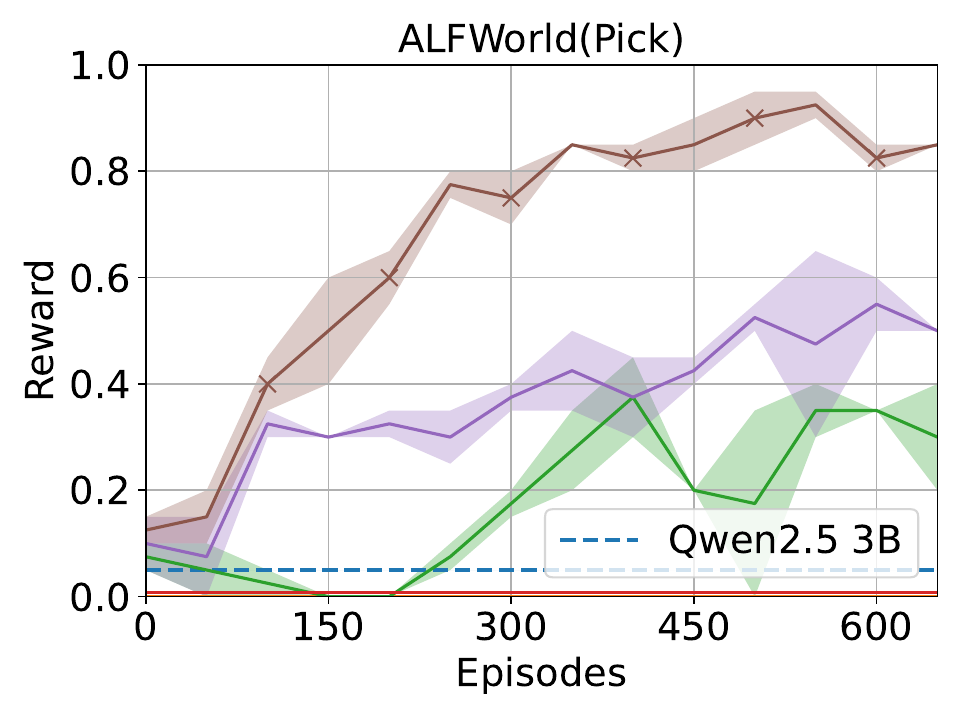}}
{\includegraphics[width=0.325\linewidth ]{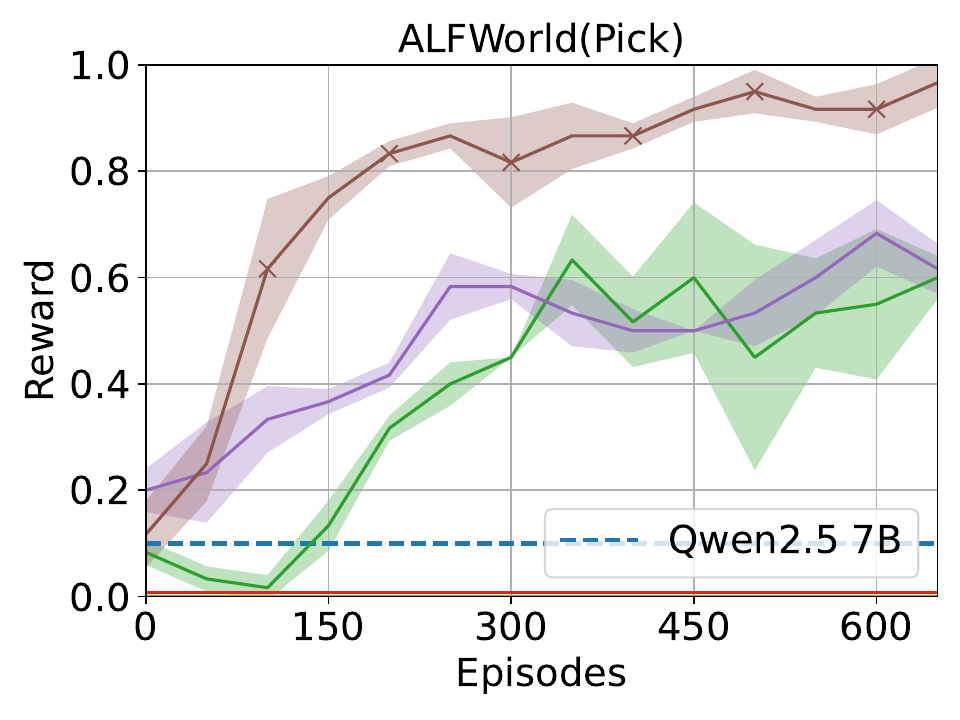}}
{\includegraphics[width=0.325\linewidth ]{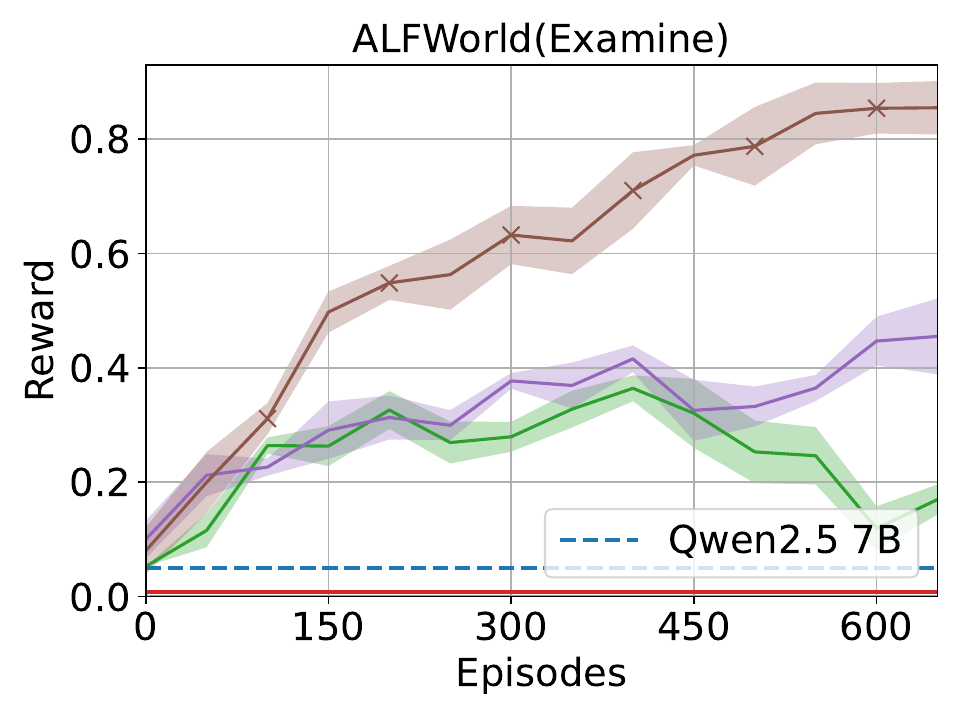}}
\vspace{-4mm}
    \caption{Results of comparison with baselines. We plot the mean and standard error of the cumulative reward. The dashed line represents directly prompting the LLM prior to generating actions given state information, with the corresponding LLM version specified. The `$\times$' markers for $\text{Mem-EM}_\text{w/ tune}$ indicate the time steps when the LLM prior is fine-tuned.}
\label{fig:baselinesdqn}
\vspace{-6mm}
\end{figure*}
\subsection{Baselines}
We compare against the following baselines:
\newline\textbf{LLM prior}: We first assess the ability of LLMs to solve decision-making tasks in a zero-shot manner, without relying on deliberately designed prompts.
\newline\textbf{DQN-based methods}: We compare against \textbf{DQN} and its variant \textbf{DQN-prior} \cite{yanefficient}, which performs deep Q-learning within a narrowed sub-action space generated by LLM priors.
\newline Our memory-driven algorithms: \textbf{Mem-Q} (Algorithm \ref{alg:train-proposal}) leverages language models' embedding capabilities and the kernel-based value estimation to perform non-parametric Q-learning.
\newline\textbf{Mem-EM} (Algorithm \ref{alg:Mem-EM}) formulates decision-making as an EM procedure that integrates memory-driven Q-value estimation with LLM prior refinement. We consider two variants: $\textbf{Mem-EM}_{\text{w/ tune}}$ fine-tunes the LLM prior using examples from the memory table at regular intervals, and $\textbf{Mem-EM}_{\text{w/o tune}}$ uses the fixed LLM prior without fine-tuning.

\textbf{Experimental Settings} We use Qwen2.5-3B \citep{team2024qwen2} or Qwen2.5-7B as the LLM prior for our main experiments. The retrieval size $M$ is set to 20, and the number of action candidates $K$ is set to 5 or 10. The fixed BERT-base \citep{devlin2019bert} model is used to obtain semantic representations of state–action pairs. Detailed hyperparameter settings are provided in the Appendix \ref{ablation:exp}.

\subsection{Results}

\textbf{Comparison with Baselines.} The comparison results of baselines are shown in Figure \ref{fig:baselinesdqn}. 
These results demonstrate that $\text{Mem-EM}_{\text{w/ tune}}$ outperforms all other baselines, remarkably achieving over 40\% improvement on complex ALFWorld environments. We observe that memory-based Mem-Q significantly outperforms DQN on Overcooked, and memory-based $\text{Mem-EM}_{\text{w/o tune}}$ achieves comparable or better performance than DQN-prior, demonstrating that the memory-driven approaches provide satisfactory value estimation and improve sample efficiency. However, for the complex ALFWorld, where both state and action spaces are extremely large, neither DQN nor Mem-Q alone can solve the task, highlighting the necessity of incorporating LLM priors.

By leveraging LLM priors to generate valuable action candidates, the baselines DQN-Prior and $\text{Mem-EM}_{\text{w/o tune}}$ consistently outperform their vanilla counterparts (DQN and Mem-Q) that explore the full original action space. Nevertheless, since pretrained LLMs lack task-specific knowledge, fixed LLM priors inherently limit performance and may even degrade it. For example, in the Overcooked (Salad) task with Qwen2.5-3B and $K=10$ candidate actions, incorporating the prior actually degrades performance. This suggests that Qwen2.5-3B cannot construct a reliable sub-action space that consistently includes the optimal action for each state, due to its insufficient domain knowledge and decision-making capability.

Importantly, results show that $\text{Mem-EM}_{\text{w/ tune}}$ substantially outperforms $\text{Mem-EM}_{\text{w/o tune}}$ on ALFWorld and Overcooked (Salad) with the 3B model. This demonstrates that our memory-driven policy optimization effectively integrates domain-specific knowledge into the LLM prior, thereby improving its decision-making ability. Furthermore, as illustrated in the ALFWorld experiments, the EM-based framework $\text{Mem-EM}_{\text{w/ tune}}$ requires only six time LLM tuning throughout training, with LoRA \cite{hulora} used for parameter-efficient fine-tuning. This keeps the computational overhead tolerable while yielding significant performance improvements.

\textbf{Generalization Ability.} The generalization ability evaluation results on ALFWorld(Pick) are shown in Table \ref{tab:generalization}. We evaluate the generalization ability of the finetuned LLM policy and the Q-value estimators including the Q-network in and the memory-driven Q-estimator defined in \eq \ref{mem_Q}. In general, we consider the components below and their combinations: the pretrained Qwen2.5-7B, denoted as $\textbf{LLM}$; the Q network trained with DQN-Prior, denoted as $\textbf{Q}_\textbf{DQN-Prior}$; the fine-tuned LLM following $\text{Mem-EM}_{\text{w/ tune}}$, denoted as $\textbf{LLM}_{\textbf{Mem-EM}_{\textbf{w/ tune}}}$; and the Q estimator of $\text{Mem-EM}_{\text{w/ tune}}$, denoted as $\textbf{Q}_{\textbf{Mem-EM}_{\textbf{w/ tune}}}$. The results show that while all methods perform well on seen tasks, the Q-estimators achieve only modest improvements over the pretrained LLM on unseen tasks. This exposes the generalization limitations of Q-estimators, which are constrained by the BERT-base representations despite being effective on seen tasks. In contrast, LLM fine-tuning shows superior generalization ability on unseen tasks, demonstrating the necessity of LLM prior refinement. Combining the fine-tuned LLM with the memory-based Q-estimator further improves performance, achieving a over 75\% performance gain than the pretrained LLM. 


\begin{table}[!t]
\vspace{-5mm}
\centering
    \caption{Results on the generalization ability of ALFWorld(Pick). All trainable models are trained with $K=5$ action candidates, and their performance is evaluated using different values of $K$.}
\resizebox{.87\linewidth}{!}{
\begin{tabular}{l|cccc|cccc}
\hline
\multirow{2}{*}{Baseline} & \multicolumn{4}{c|}{Unseen Tasks} & \multicolumn{4}{c}{Seen Tasks} \\
& K=5 & K=10 & K=15 & K=20 & K=5 & K=10 & K=15 & K=20 \\
\hline
LLM &\multicolumn{4}{c|}{0.19} & \multicolumn{4}{c}{0.1}\\
LLM + Q$_{\text{DQN-Prior}}$ & 0.16 & 0.19 & 0.14 & $\mathbf{0.22}$ & 0.6 & $\mathbf{0.70}$ & $\mathbf{0.70}$ & 0.65 \\
LLM + Q$_{\text{Mem-EM}_\text{w/{tune}}}$ & 0.22 & ${0.27}$ & $\mathbf{0.35}$ & 0.22 & 0.65 & 0.80 & $\mathbf{0.85}$ & 0.65 \\
LLM$_{\text{Mem-EM}_\text{w/{tune}}}$ & \multicolumn{4}{c|}{0.59} & \multicolumn{4}{c}{0.85} \\
LLM$_{\text{Mem-EM}_\text{w/{tune}}}$+Q$_{\text{DQN-Prior}}$ & $\mathbf{0.59}$ & 0.54 & 0.46 & 0.46 & $\mathbf{0.90}$ & 0.85 & $\mathbf{0.90}$ & 0.75 \\
LLM$_{\text{Mem-EM}_\text{w/{tune}}}$+Q$_{\text{Mem-EM}_\text{w/{tune}}}$ & $\mathbf{0.81}$ & 0.65 & 0.62 & 0.68 & 0.95 & 0.95 & $\mathbf{1.00}$ & 0.95 \\
\hline
\end{tabular}}
\vspace{-4mm}
\label{tab:generalization}
\end{table}

\begin{figure*}[!t]
\centering
\vspace{-1mm}
\subfigure[Action candidates]
{\label{fig:actionk}\includegraphics[width=0.325\linewidth]{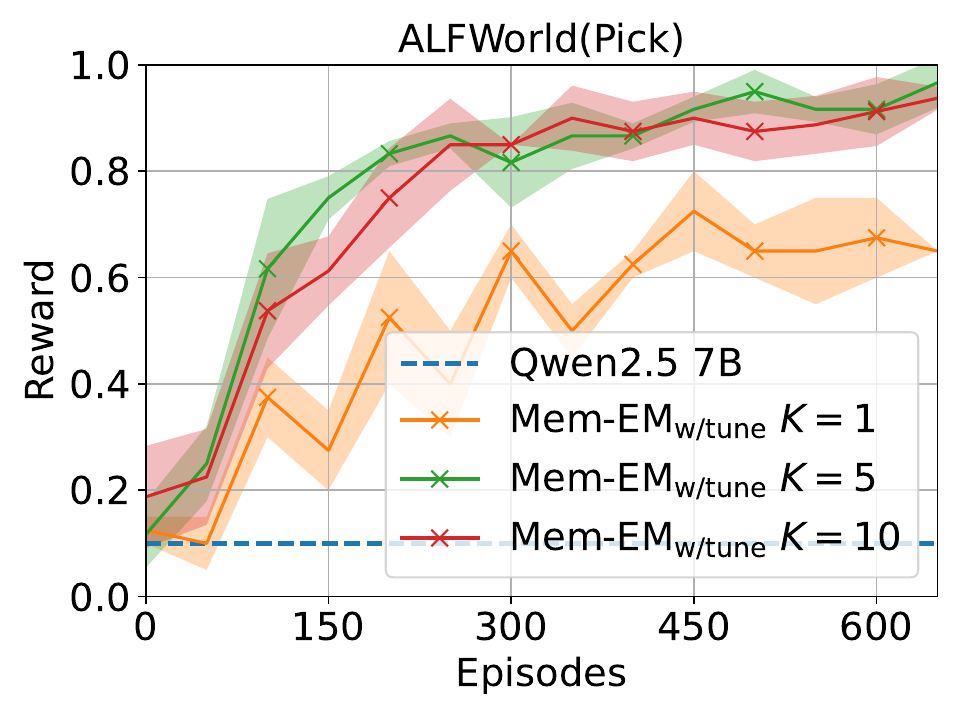}}
\subfigure[Tuning interval]
{\label{fig:tuning interval}\includegraphics[width=0.325\linewidth]{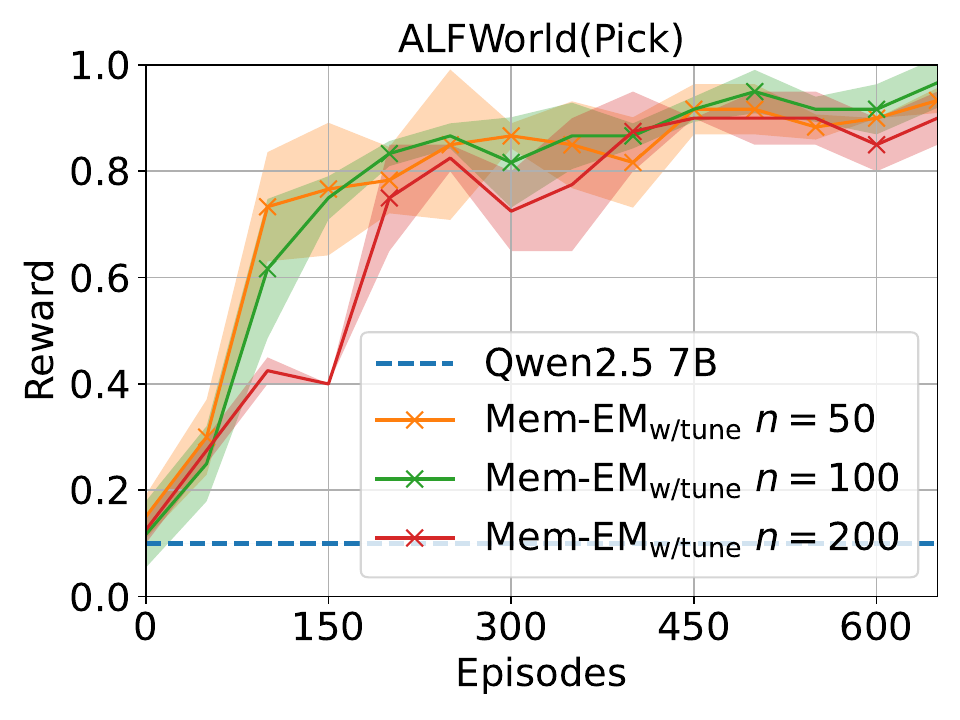}}
\subfigure[Memory capacity]
{\label{fig:memsize}\includegraphics[width=0.325\linewidth]{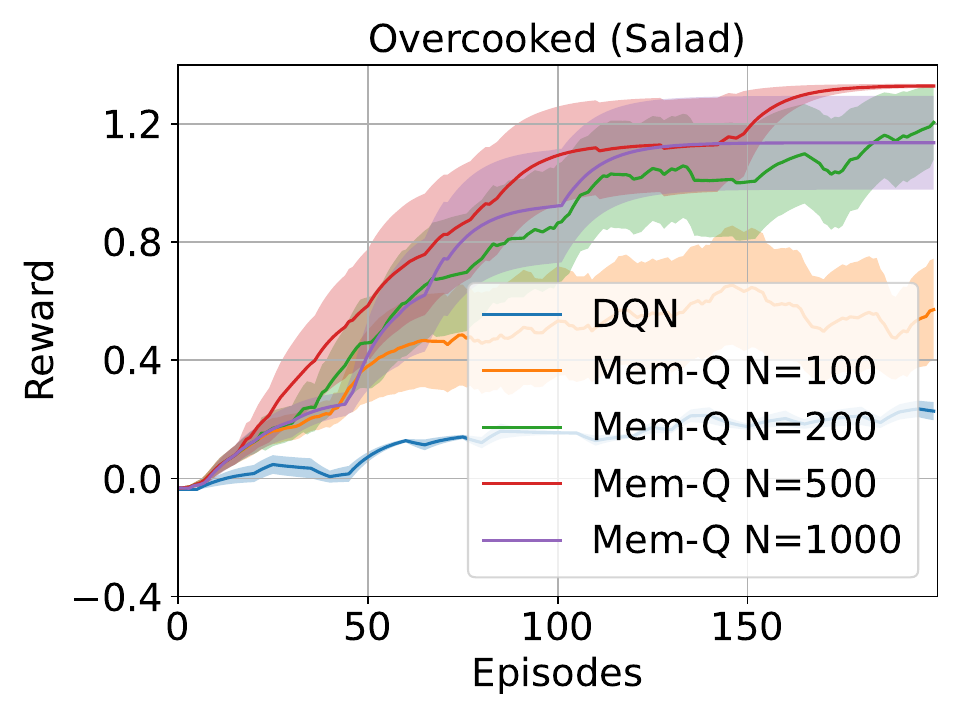}}
\vspace{-3mm}
\caption{Ablation study results. (a) Effect of the number of action candidates $K$ generated by the LLM. The `$\times$' markers indicate the time steps when the LLM-prior is updated. (b) Impact of the LLM fine-tuning interval, where $n$ denotes that the LLM policy is fine-tuned every $n$ episodes. (c) Influence of the memory table capacity $N$, where at most $N$ $(s,a)$ pairs are stored, with the least-recently-used (LRU) strategy applied for replacement.}
\label{fig:ablation}
\vspace{-3mm}
\end{figure*}

\subsection{Ablation Studies}
The ablation study results are shown in \fig \ref{fig:ablation}, where we analyze the following aspects:
\newline\textbf{Effect of the number of action candidates.}  
Figure \ref{fig:actionk} illustrates the impact of the number of action candidates $K$ for $\text{Mem-EM}_\text{w/ tune}$. Our method with $K=5$ or $10$ significantly outperforms the setting with $K=1$, indicating that it benefits from both (1) LLM fine-tuning that incorporates domain-specific knowledge stored in the memory table, and (2) Q-value guided posterior action sampling. It is worth noting that the case of $\text{Mem-EM}_\text{w/ tune}$ with $K=1$ resembles an actor–critic RL setting, where the action policy samples trajectories and the critic model is trained with environment returns to guide the policy learning. These results highlight the superiority of our EM-based training framework over traditional actor–critic RL framework. This superiority is achieved because, in our approach, the policy is treated as a prior, while sampling is performed from the action posterior and guided by memory-driven value estimation, jointly resulting in higher sample efficiency.
\newline
\textbf{Effect of the LLM prior update interval.}  
Fig.~\ref{fig:tuning interval} examines the effect of the fine-tuning interval $n$, where the LLM prior is updated every $n$ episodes. Results show that $\text{Mem-EM}_\text{w/ tune}$ is robust to the update interval and does not require frequent fine-tuning to achieve strong performance.
\newline
\textbf{Effect of memory capacity.}  
\fig \ref{fig:memsize} shows the effect of memory table capacity $N$ on Overcooked (Salad). In this environment, the total number of possible $(s,a)$ pairs is approximately $1000$. Remarkably, {Mem-Q} with only $N=100$ entries already outperforms DQN, and configurations with $N=200$ or $500$ achieve performance comparable to $N=1000$, which stores all possible pairs. This demonstrates that, by simply using a least-recently-used (LRU) replacement strategy to retain crucial $(s,a)$ pairs, memory-driven Q estimation remains robust to memory capacity.

\section {Related Work}
\textbf{LLM Priors in Decision-Making}
Recent works leverage LLMs to enhance sequential decision-making (SDM) in three main ways: action generation, value estimation, and reward function design. First, LLMs can act as the action policy, generating satisfactory actions either through deliberate prompting \cite{yao2022react, shinn2024reflexion} or RL-based fine-tuning \cite{carta2023grounding, tantrue}. Second, LLMs can serve as the value function to guide the search. Examples include reasoning-path search \cite{wang2022self, yao2023tree} and Monte Carlo Tree Search guided by LLM evaluations \cite{hao2023reasoning, wan2024alphazero}. In addition, LLMs can be fine-tuned to act as process or outcome reward models with detailed explanations \cite{lightman2023let, mcaleese2024llm, wang2024math}. Third, LLMs are used to generate reward signals for RL, either directly by prompting with historical interactions \cite{kwonreward} or by producing executable reward code for continuous-control tasks \cite{yu2023language, maeureka}.
\newline\textbf{Memory-based Decision-Making}
Episodic Control (EC) \cite{blundell2016model} and its extensions \cite{pritzel2017neural, li2023neural} represent a classic family of memory-based methods. These approaches maintain $|\mathcal{A}|$ separate memory tables, one for each feasible action, and apply kernel-based estimation over state representations. Yet, EC relies solely on state similarity and ignores semantic relationships among actions. Off-policy methods such as DQN \cite{mnih2013playing} and SAC \cite{haarnoja2018soft} can also be viewed as memory-based RL, with the replay buffer serving as memory. \cite{yan2024efficient} further combine DQN with LLM embeddings, but their approach compresses stored experiences into Q-networks that map high-dimensional embeddings to scalar values, potentially discarding semantic information and limiting sample efficiency. More recently, memory-based methods have been integrated with LLMs under the retrieval-augmented generation (RAG) paradigm \cite{zhou2025agentfly, wang2024biorag}. These approaches retrieve relevant cases to enhance LLM outputs via in-context learning \cite{han2023explaining}, but they primarily focus on one-step tasks such as question answering \cite{wiratunga2024cbr} or high-level planning \cite{zhou2025agentfly}. By contrast, our work applies retrieval techniques to sequential decision-making, and explicitly incorporates domain-specific memory to refine LLM-based policies.

\section{Conclusions}
In this work, we propose a memory-driven self-improvement framework for decision-making tasks. The framework consists of two mutually reinforcing components: memory-driven value estimation and memory-driven LLM prior refinement. The memory-driven value estimation approach maintains historical interactions and their Q-values, performing non-parametric value estimation through retrieval of similar representations. Building on the EM formulation, we design a practical and stable memory-driven LLM prior refinement algorithm, which adapts task-specific knowledge into the LLM prior by learning from the memory table. The explicit use of memory in these two components encourages efficient exploration. Experimental results show that our EM-based self-improvement framework delivers substantial performance gains while avoiding extensive fine-tuning. In this work, we focus on text-based decision-making with discrete but enumerable action spaces. For future research, we plan to extend our framework to handle scenarios with free-form or infinite action spaces and to incorporate vision–language models, thereby enabling broader applications.
\newpage
\balance
\bibliography{iclr2026_conference}
\bibliographystyle{iclr2026_conference}
\include{appendix}
\end{document}

%% file: math_commands.tex
\usepackage{amsmath,amsfonts,bm}

\def\eqref#1{equation~\ref{#1}}

\def\1{\bm{1}}

\DeclareMathAlphabet{\mathsfit}{\encodingdefault}{\sfdefault}{m}{sl}
\SetMathAlphabet{\mathsfit}{bold}{\encodingdefault}{\sfdefault}{bx}{n}

\newcommand{\E}{\mathbb{E}}

\DeclareMathOperator*{\argmax}{arg\,max}
\DeclareMathOperator*{\argmin}{arg\,min}

%% file: appendix.tex
\section{Additional Explaination on Memory-driven value estimation}
\begin{figure*}[th!]
\centering
{\includegraphics[width=0.425\linewidth]{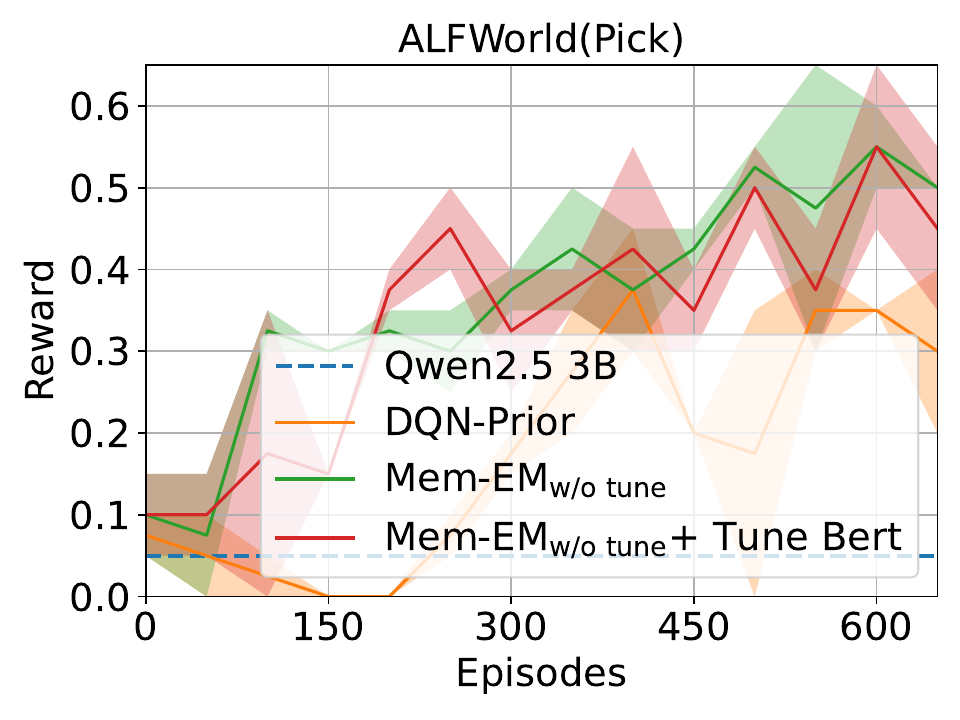}}
{\includegraphics[width=0.425\linewidth]{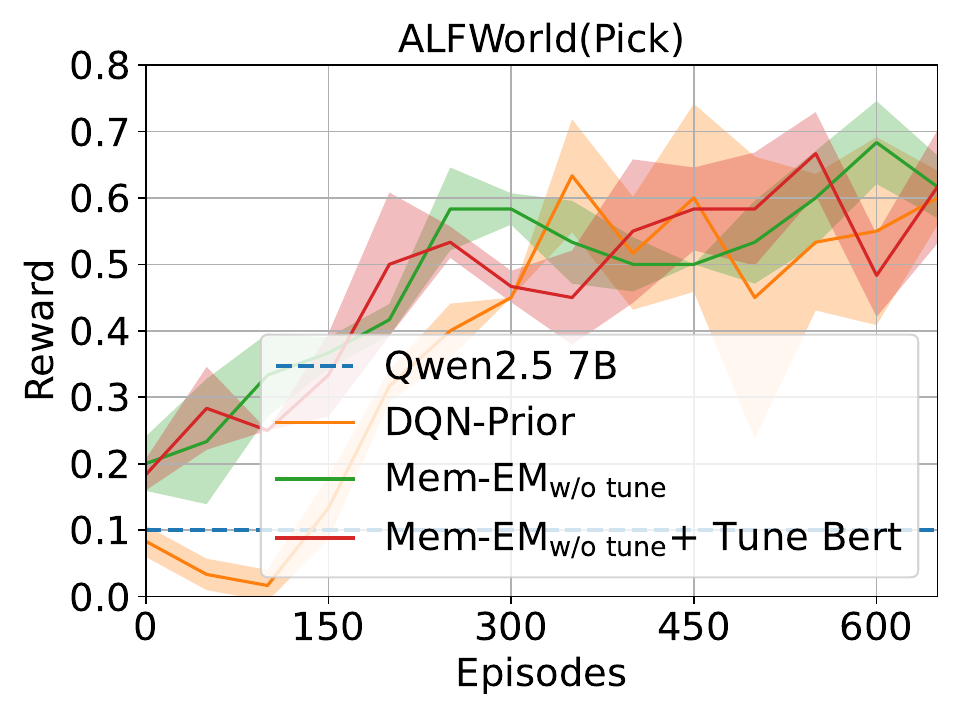}}

\caption{Ablation study on finetuning the embedding model bert. }
\label{fig:bertfine}
\end{figure*}

In the main paper, we presented memory-driven Q-learning estimation without training. Here, we explore an alternative approach that involves tuning the embedding function. As described above, we use the BERT-base model as the embedding function $f_\theta$, which can be further fine-tuned by minimizing the distance between the predicted Q-value and the Monte Carlo estimate:

$$
\ell = -(\hat{Q}_\theta(s_t, a) - y_t)^2,
$$

where $y_t = \sum_{i=t}^T \gamma^{i-t} r_t$. The results of fine-tuning the BERT embedding model are shown in Figure \ref{fig:bertfine}. We observe that our model, both with and without fine-tuning, outperforms DQN-Prior. All three baselines use the LLM prior to narrow the action search space, and fine-tuning versus keeping the BERT embedding model fixed yields similar performance.

\section{Additional Explanation of Probabilistic Inference}
In this section, we briefly review the Expectation–Maximization (EM) algorithm.
We start by introducing the evidence lower bound (ELBO) for the log-likelihood:
\begin{align}
\log p_\theta(\mathcal{O}=1|s)
=&\sum_a\log p_\theta(\mathcal{O}=1|s,a)p_\theta(a|s)\\
=&\sum_a \log \left[q_\phi(a|s)\frac{p_\theta(\mathcal{O}=1|s,a)p_\theta(a|s)}{q_\phi(a|s)}\right]\\
\geq& \mathbb{E}_{q_\phi(a|s)}[\log p_\theta(\mathcal{O}=1|s,a)]-D_\text{KL}(q_\phi(a|s)\|p_\theta(a|s)) \triangleq \mathrm{ELBO}(\phi, \theta)
\end{align}
In variational inference, maximizing the likelihood can be reformulated as the following two-step optimization:
\begin{align}
    \phi_{k+1} &\leftarrow \argmax_\phi \mathrm{ELBO}(\phi, \theta_k) \label{eq:first-step-vi} \\
    \theta_{k+1} &\leftarrow \argmax_\theta \mathrm{ELBO}(\phi_{k+1}, \theta).
\end{align}
It can be shown that when $q_{\phi_{k+1}} = p_{\theta_k}(a|s,\mathcal{O}=1)$, the $\mathrm{ELBO}$ attains its maximum value, which coincides with the true log-likelihood:
\begin{align}
&\mathbb{E}_{p_\theta(a|s,\mathcal{O}=1)}[\log p_\theta(\mathcal{O}=1|s,a)]-D_\text{KL}(p_\theta(a|s,\mathcal{O}=1)\|p_\theta(a|s))\\
=&\mathbb{E}_{p_\theta(a|s,\mathcal{O}=1)}\log\left[\frac{p_\theta(a,\mathcal{O}=1|s)}{p_\theta(a|s,\mathcal{O}=1)}\right]\\
=&\mathbb{E}_{p_\theta(a|s,\mathcal{O}=1)}\log\left[\frac{p_\theta(a|s,\mathcal{O}=1)p_\theta(\mathcal{O}=1|s)}{p_\theta(a|s,\mathcal{O}=1)}\right]\\
=&\mathbb{E}_{p_\theta(a|s,\mathcal{O}=1)}[\log p_\theta(\mathcal{O}=1 | s)]\\
=& \log p_\theta(\mathcal{O}=1 | s).
\end{align}
Thus, EM maximizes the following objective:
\begin{align}
    \theta_{k+1} = \argmax_{\theta} \mathbb{E}_{p_{\theta_{k}}(a|s,\mathcal{O}=1)}[\log p_\theta(\mathcal{O}=1|s,a)]-D_\text{KL}(p_{\theta_{k}}(a|s,\mathcal{O}=1)\|p_\theta(a|s),
\end{align}
which is equivalent to
\begin{align}
    \theta_{k+1} = \argmax_{\theta} \E_{p_{\theta_k}(a,s | \mathcal{O}=1)} \left[ \log p_\theta(\mathcal{O}=1|s,a) + \log p_{\theta} (a|s) \right].
\end{align}
Although this optimization is generally intractable, in the E-step one can approximate the expectation using Monte Carlo estimation, followed by the M-step to update the parameter $\theta$.
Finally, since the ELBO lower bounds the log-likelihood, each EM iteration guarantees monotonic improvement $ \log p_{\theta_{k+1}} (\mathcal{O}=1 | s) \ge \log p_{\theta_{k}} (\mathcal{O}=1 | s)$.

\section{Detailed Experimental Settings}
\label{ablation:exp}
\subsection{Environments}
{}{Examples of observations and admissible actions for ALFWorld and Overcooked are shown below:}
\begin{tcolorbox}[title={For ALFWorld(Pick)}]
    
\textbf{Observation:}
Task: Your task is to: put some knife on sidetable.
Current observation:You open the drawer 2. The drawer 2 is open. In it, you see nothing.. 

\textbf{Admissible actions} You are allowed to take the following actions: close drawer 2, examine drawer 2, go to cabinet 1, go to cabinet 2, go to cabinet 3, go to cabinet 4, go to coffeemachine 1, go to countertop 1, go to drawer 1, go to drawer 3, go to drawer 4, go to drawer 5, go to drawer 6, go to drawer 7, go to drawer 8, go to fridge 1, go to garbagecan 1, go to microwave 1, go to sidetable 1, go to sinkbasin 1, go to stoveburner 1, go to stoveburner 2, go to stoveburner 3, go to stoveburner 4, go to toaster 1, inventory, look.
\end{tcolorbox}

\begin{tcolorbox}[title={For Overcooked(Salad)}]
    
\textbf{Observation:}
There are two fixed cutting boards in the room. You notice a tomato and an onion on the different tables. Currently you are carrying an unchopped lettuce in hand.
To serve the dish of a bowl only containing chopped tomato and lettuce, what action should you take next? 

\textbf{Admissible actions} You are allowed to take the following actions: 'pick up the tomato', 'pick up the lettuce', 'pick up the onion', 'take the empty bowl', 'put the lettuce on the first cutting board', 'put the lettuce on the second cutting board', 'serve the dish', 'chop nothing
\end{tcolorbox}

\subsection{LLM prior implementation}
\label{llm_prior_setting}

Following \cite{yan2024efficient}, we use the LLM prior to generate $K$ action candidates by sampling $K$ free-form outputs in parallel from the LLM, given the current state and all admissible actions. These outputs are then mapped to executable actions via a simple rule-based projection $\mathcal{P}$. Mathematically, this can be described as:
$
a \sim p_{\text{LLM}}(\cdot \mid s_t) \longleftrightarrow o \sim \text{LLM}(\cdot \mid s_t), \ a = \mathcal{P}(o),
$
 where $\mathcal{P}$ is a rule-based mapping that selects the most frequently occurring action in the LLM output $o$.


An Example of LLM output and rule-based mapping is given as:
\begin{tcolorbox}[title={For ALFWorld(Pick) from Qwen2.5 7B}]
    
\textbf{Input:}
Task: Your task is to: put some alarmclock on desk.
Current observation:You turn on the desklamp 1.. You are allowed to take the following actions: examine desk 1, examine keychain 3, go to bed 1, go to drawer 1, go to drawer 2, go to drawer 3, go to drawer 4, go to drawer 5, go to dresser 1, go to garbagecan 1, inventory, look, put keychain 3 in/on desk 1, use desklamp 1.

Please select an action from the admissible actions. Please just output the selected action:

\textbf{LLM Output to Action:} 
1: put keychain 3 in/on desk $\rightarrow$ put keychain 3 in/on desk

2: examine keychain 3 $\rightarrow$ examine keychain 3

3: I will choose "put keychain 3 in/on desk" $\rightarrow$ put keychain 3 in/on desk

4: You just picked up the desklamp 1 $\rightarrow$ Randomly select an action, since no feasible action appears in the output  

5: use desklamp 1 $\rightarrow$ use desklamp 1  

We map the LLM's output to an executable action using a simple rule-based method, extracting the executable actions directly from the LLM's output.
\end{tcolorbox}
\subsection{Computational Resources and Hyperparameters}
Our experiments are conducted on a single machine equipped with eight 48 GB A6000 GPUs, using PyTorch 2.1 with CUDA 12.4. Tables \ref{tab:tomato}, \ref{tab:salad}, \ref{tab:pick}, and \ref{tab:exam} report the main hyperparameters of our algorithms. For the likelihood, we approximate it as $
p(\mathcal{O}=1 \mid s,a) \propto \exp(Q(s,a)/\tau),
$ where the hyperparameter $\tau$ is used in two contexts: (1) value-function-guided action posterior selection, and (2) reweighting the log-likelihood during LLM prior fine-tuning, as shown in Eq.~\ref{priorlearn}. In practice, we tune these two parameters separately and denote them as $\tau_1$ and $\tau_2$, respectively. The horizon and memory capacity of environments are shown in Table \ref{tab:horizon}.
\begin{table}[h!]
    \caption{The hyperparameters on Overcooked(Tomato
    )}
    \centering
    \begin{tabular}{ccccccccc}
    \hline
   Baselines & Learning Rate & Epochs & Batch Size & Update Frequency& $\tau_1$  &$\tau_2$& K\\
    \hline
        Mem-EM$_\text{w/ tune}$ & 5e-4& 3&16&  10& 0.1&0.5 &5\\
    \hline
        \end{tabular}
    \label{tab:tomato}
\end{table}
\begin{table}[h!]
    \caption{The hyperparameters on Overcooked(Salad)}
    \centering
    \begin{tabular}{ccccccccc}
    \hline
   Baselines & Learning Rate & Epochs & Batch Size & Update Frequency& $\tau_1$  &$\tau_2$& K\\
    \hline
        Mem-EM$_\text{w/ tune}$ & 5e-4& 3&16&  10& 0.1&0.5 &10\\
    \hline
        \end{tabular}
    \label{tab:salad}
\end{table}
\begin{table}[h!]
    \caption{The hyperparameters on ALFWorld(Pick)}
    \centering
    \begin{tabular}{ccccccccc}
    \hline
    Baselines & Learning Rate & Epochs & Batch Size & Update Frequency& $\tau_1$  &$\tau_2$& K\\
    \hline
        Mem-EM$_\text{w/ tune}$ & 5e-4& 3&16&  100& 0.1&0.2 &5\\
    \hline
        \end{tabular}
    \label{tab:pick}
\end{table}
\begin{table}[h!]
    \caption{The hyperparameters on ALFWorld(Examine)}
    \centering
    \begin{tabular}{ccccccccc}
    \hline
   Baselines & Learning Rate & Epochs & Batch Size & Update Frequency& $\tau_1$  &$\tau_2$& K\\
    \hline
        Mem-EM$_\text{w/ tune}$ & 5e-4& 3&16&  100& 0.1&0.5 &10\\
    \hline
        \end{tabular}
    \label{tab:exam}
\end{table}

\begin{table}[]
    \centering
    \begin{tabular}{c|ccccc}
    \hline
         & ALFWorld(Pick) &ALFWolrd(Examine) &Cook(Tomato)&Cook(Salad)  \\
         \hline
         Horizon& 60&60&15&30 \\
         Memory Capacity&100&1000&15000&15000\\
         \hline
    \end{tabular}
    \caption{Maximium horizon and memory capacity of environments. }
    \label{tab:horizon}
\end{table}

%% file: main.bbl
\begin{thebibliography}{49}
\providecommand{\natexlab}[1]{#1}
\providecommand{\url}[1]{\texttt{#1}}
\expandafter\ifx\csname urlstyle\endcsname\relax
  \providecommand{\doi}[1]{doi: #1}\else
  \providecommand{\doi}{doi: \begingroup \urlstyle{rm}\Url}\fi

\bibitem[Abdolmaleki et~al.(2018)Abdolmaleki, Springenberg, Tassa, Munos, Heess, and Riedmiller]{abdolmaleki2018maximum}
Abbas Abdolmaleki, Jost~Tobias Springenberg, Yuval Tassa, Remi Munos, Nicolas Heess, and Martin Riedmiller.
\newblock Maximum a posteriori policy optimisation.
\newblock \emph{arXiv preprint arXiv:1806.06920}, 2018.

\bibitem[Blundell et~al.(2016)Blundell, Uria, Pritzel, Li, Ruderman, Leibo, Rae, Wierstra, and Hassabis]{blundell2016model}
Charles Blundell, Benigno Uria, Alexander Pritzel, Yazhe Li, Avraham Ruderman, Joel~Z Leibo, Jack Rae, Daan Wierstra, and Demis Hassabis.
\newblock Model-free episodic control.
\newblock \emph{arXiv preprint arXiv:1606.04460}, 2016.

\bibitem[Brunke et~al.(2022)Brunke, Greeff, Hall, Yuan, Zhou, Panerati, and Schoellig]{brunke2022safe}
Lukas Brunke, Melissa Greeff, Adam~W Hall, Zhaocong Yuan, Siqi Zhou, Jacopo Panerati, and Angela~P Schoellig.
\newblock Safe learning in robotics: From learning-based control to safe reinforcement learning.
\newblock \emph{Annual Review of Control, Robotics, and Autonomous Systems}, 5\penalty0 (1):\penalty0 411--444, 2022.

\bibitem[Carta et~al.(2023)Carta, Romac, Wolf, Lamprier, Sigaud, and Oudeyer]{carta2023grounding}
Thomas Carta, Cl{\'e}ment Romac, Thomas Wolf, Sylvain Lamprier, Olivier Sigaud, and Pierre-Yves Oudeyer.
\newblock Grounding large language models in interactive environments with online reinforcement learning.
\newblock In \emph{International Conference on Machine Learning}, pp.\  3676--3713. PMLR, 2023.

\bibitem[Chen et~al.(2023)Chen, Diao, Chen, Yao, Piao, Sun, Yang, Goebel, Jiang, and Chang]{chen2023sufficiency}
Xing Chen, Dongcui Diao, Hechang Chen, Hengshuai Yao, Haiyin Piao, Zhixiao Sun, Zhiwei Yang, Randy Goebel, Bei Jiang, and Yi~Chang.
\newblock The sufficiency of off-policyness and soft clipping: Ppo is still insufficient according to an off-policy measure.
\newblock In \emph{Proceedings of the AAAI Conference on Artificial Intelligence}, volume~37, pp.\  7078--7086, 2023.

\bibitem[Christianos et~al.(2023)Christianos, Papoudakis, Zimmer, Coste, Wu, Chen, Khandelwal, Doran, Feng, Liu, Xiong, Luo, Hao, Shao, Bou-Ammar, and Wang]{christianos2023panguagent}
Filippos Christianos, Georgios Papoudakis, Matthieu Zimmer, Thomas Coste, Zhihao Wu, Jingxuan Chen, Khyati Khandelwal, James Doran, Xidong Feng, Jiacheng Liu, Zheng Xiong, Yicheng Luo, Jianye Hao, Kun Shao, Haitham Bou-Ammar, and Jun Wang.
\newblock Pangu-agent: A fine-tunable generalist agent with structured reasoning, 2023.

\bibitem[Devlin et~al.(2019)Devlin, Chang, Lee, and Toutanova]{devlin2019bert}
Jacob Devlin, Ming-Wei Chang, Kenton Lee, and Kristina Toutanova.
\newblock Bert: Pre-training of deep bidirectional transformers for language understanding.
\newblock In \emph{Proceedings of the 2019 conference of the North American chapter of the association for computational linguistics: human language technologies, volume 1 (long and short papers)}, pp.\  4171--4186, 2019.

\bibitem[Granter et~al.(2017)Granter, Beck, and Papke~Jr]{granter2017alphago}
Scott~R Granter, Andrew~H Beck, and David~J Papke~Jr.
\newblock Alphago, deep learning, and the future of the human microscopist.
\newblock \emph{Archives of pathology \& laboratory medicine}, 141\penalty0 (5):\penalty0 619--621, 2017.

\bibitem[Haarnoja et~al.(2018)Haarnoja, Zhou, Abbeel, and Levine]{haarnoja2018soft}
Tuomas Haarnoja, Aurick Zhou, Pieter Abbeel, and Sergey Levine.
\newblock Soft actor-critic: Off-policy maximum entropy deep reinforcement learning with a stochastic actor.
\newblock In \emph{International Conference on Machine Learning (ICML)}, pp.\  1861--1870. PMLR, 2018.

\bibitem[Han et~al.(2023)Han, Wang, Zhao, and Ji]{han2023explaining}
Chi Han, Ziqi Wang, Han Zhao, and Heng Ji.
\newblock Explaining emergent in-context learning as kernel regression.
\newblock \emph{arXiv preprint arXiv:2305.12766}, 2023.

\bibitem[Hao et~al.()Hao, Gu, Ma, Hong, Wang, Wang, and Hu]{hao2023reasoning}
Shibo Hao, Yi~Gu, Haodi Ma, Joshua~Jiahua Hong, Zhen Wang, Daisy~Zhe Wang, and Zhiting Hu.
\newblock Reasoning with language model is planning with world model.
\newblock In \emph{The 2023 Conference on Empirical Methods in Natural Language Processing}.

\bibitem[Hu et~al.(2021)Hu, Wallis, Allen-Zhu, Li, Wang, Wang, Chen, et~al.]{hulora}
Edward~J Hu, Phillip Wallis, Zeyuan Allen-Zhu, Yuanzhi Li, Shean Wang, Lu~Wang, Weizhu Chen, et~al.
\newblock Lora: Low-rank adaptation of large language models.
\newblock In \emph{International Conference on Learning Representations}, 2021.

\bibitem[Jannala(2025)]{jannala2025chess}
Sai~Dhanush Jannala.
\newblock \emph{Chess LLM Arena: A Framework for Evaluating Strategic Decision-Making in Large Language Models}.
\newblock PhD thesis, Dublin, National College of Ireland, 2025.

\bibitem[Jin et~al.(2024)Jin, Wang, Du, Fang, Zhang, and Wang]{jin2024learning}
Xuanfa Jin, Ziyan Wang, Yali Du, Meng Fang, Haifeng Zhang, and Jun Wang.
\newblock Learning to discuss strategically: A case study on one night ultimate werewolf.
\newblock \emph{Advances in Neural Information Processing Systems}, 37:\penalty0 77060--77097, 2024.

\bibitem[Klissarov et~al.(2023)Klissarov, D'Oro, Sodhani, Raileanu, Bacon, Vincent, Zhang, and Henaff]{klissarov2023motif}
Martin Klissarov, Pierluca D'Oro, Shagun Sodhani, Roberta Raileanu, Pierre-Luc Bacon, Pascal Vincent, Amy Zhang, and Mikael Henaff.
\newblock Motif: Intrinsic motivation from artificial intelligence feedback.
\newblock \emph{arXiv preprint arXiv:2310.00166}, 2023.

\bibitem[Kwon et~al.(2023)Kwon, Xie, Bullard, and Sadigh]{kwonreward}
Minae Kwon, Sang~Michael Xie, Kalesha Bullard, and Dorsa Sadigh.
\newblock Reward design with language models.
\newblock In \emph{The Eleventh International Conference on Learning Representations}, 2023.

\bibitem[Levine(2018)]{levine2018reinforcement}
Sergey Levine.
\newblock Reinforcement learning and control as probabilistic inference: Tutorial and review.
\newblock \emph{arXiv preprint arXiv:1805.00909}, 2018.

\bibitem[Li et~al.(2024)Li, Jelassi, Zhang, Kakade, Wattenberg, and Brandfonbrener]{li2024q}
Kenneth Li, Samy Jelassi, Hugh Zhang, Sham~M Kakade, Martin Wattenberg, and David Brandfonbrener.
\newblock Q-probe: A lightweight approach to reward maximization for language models.
\newblock In \emph{International Conference on Machine Learning}, pp.\  27955--27968. PMLR, 2024.

\bibitem[Li et~al.(2023)Li, Zhu, Hu, Xie, Ma, Zheng, Song, Chen, and Zhao]{li2023neural}
Zhuo Li, Derui Zhu, Yujing Hu, Xiaofei Xie, Lei Ma, Yan Zheng, Yan Song, Yingfeng Chen, and Jianjun Zhao.
\newblock Neural episodic control with state abstraction.
\newblock \emph{arXiv preprint arXiv:2301.11490}, 2023.

\bibitem[Li et~al.(2019)Li, Kiseleva, and De~Rijke]{li2019dialogue}
Ziming Li, Julia Kiseleva, and Maarten De~Rijke.
\newblock Dialogue generation: From imitation learning to inverse reinforcement learning.
\newblock In \emph{Proceedings of the AAAI conference on artificial intelligence}, volume~33, pp.\  6722--6729, 2019.

\bibitem[Lightman et~al.(2023)Lightman, Kosaraju, Burda, Edwards, Baker, Lee, Leike, Schulman, Sutskever, and Cobbe]{lightman2023let}
Hunter Lightman, Vineet Kosaraju, Yura Burda, Harri Edwards, Bowen Baker, Teddy Lee, Jan Leike, John Schulman, Ilya Sutskever, and Karl Cobbe.
\newblock Let's verify step by step.
\newblock \emph{arXiv preprint arXiv:2305.20050}, 2023.

\bibitem[Ma et~al.(2024{\natexlab{a}})Ma, Mi, Zeng, Yan, Lin, Wu, Wang, and Zhang]{ma2024large}
Weiyu Ma, Qirui Mi, Yongcheng Zeng, Xue Yan, Runji Lin, Yuqiao Wu, Jun Wang, and Haifeng Zhang.
\newblock Large language models play starcraft ii: Benchmarks and a chain of summarization approach.
\newblock \emph{Advances in Neural Information Processing Systems}, 37:\penalty0 133386--133442, 2024{\natexlab{a}}.

\bibitem[Ma et~al.(2024{\natexlab{b}})Ma, Liang, Wang, Huang, Bastani, Jayaraman, Zhu, Fan, and Anandkumar]{maeureka}
Yecheng~Jason Ma, William Liang, Guanzhi Wang, De-An Huang, Osbert Bastani, Dinesh Jayaraman, Yuke Zhu, Linxi Fan, and Anima Anandkumar.
\newblock Eureka: Human-level reward design via coding large language models.
\newblock In \emph{The Twelfth International Conference on Learning Representations}, 2024{\natexlab{b}}.

\bibitem[McAleese et~al.(2024)McAleese, Pokorny, Uribe, Nitishinskaya, Trebacz, and Leike]{mcaleese2024llm}
Nat McAleese, Rai~Michael Pokorny, Juan Felipe~Ceron Uribe, Evgenia Nitishinskaya, Maja Trebacz, and Jan Leike.
\newblock Llm critics help catch llm bugs.
\newblock \emph{arXiv preprint arXiv:2407.00215}, 2024.

\bibitem[McTear(2022)]{mctear2022conversational}
Michael McTear.
\newblock \emph{Conversational ai: Dialogue systems, conversational agents, and chatbots}.
\newblock Springer Nature, 2022.

\bibitem[Mnih(2013)]{mnih2013playing}
Volodymyr Mnih.
\newblock Playing atari with deep reinforcement learning.
\newblock \emph{arXiv preprint arXiv:1312.5602}, 2013.

\bibitem[Naranjo et~al.(2005)Naranjo, Gonz{\'a}lez, Garc{\'\i}a, de~Pedro, and Haber]{naranjo2005power}
Jos{\'e}~Eugenio Naranjo, Carlos Gonz{\'a}lez, Ricardo Garc{\'\i}a, Teresa de~Pedro, and Rodolfo~E Haber.
\newblock Power-steering control architecture for automatic driving.
\newblock \emph{Ieee transactions on intelligent transportation systems}, 6\penalty0 (4):\penalty0 406--415, 2005.

\bibitem[Polydoros \& Nalpantidis(2017)Polydoros and Nalpantidis]{polydoros2017survey}
Athanasios~S Polydoros and Lazaros Nalpantidis.
\newblock Survey of model-based reinforcement learning: Applications on robotics.
\newblock \emph{Journal of Intelligent \& Robotic Systems}, 86\penalty0 (2):\penalty0 153--173, 2017.

\bibitem[Pritzel et~al.(2017)Pritzel, Uria, Srinivasan, Badia, Vinyals, Hassabis, Wierstra, and Blundell]{pritzel2017neural}
Alexander Pritzel, Benigno Uria, Sriram Srinivasan, Adria~Puigdomenech Badia, Oriol Vinyals, Demis Hassabis, Daan Wierstra, and Charles Blundell.
\newblock Neural episodic control.
\newblock In \emph{International conference on machine learning}, pp.\  2827--2836. PMLR, 2017.

\bibitem[Rana et~al.(2023)Rana, Xu, Tidd, Milford, and S{\"u}nderhauf]{rana2023residual}
Krishan Rana, Ming Xu, Brendan Tidd, Michael Milford, and Niko S{\"u}nderhauf.
\newblock Residual skill policies: Learning an adaptable skill-based action space for reinforcement learning for robotics.
\newblock In \emph{Conference on Robot Learning}, pp.\  2095--2104. PMLR, 2023.

\bibitem[Shinn et~al.(2024)Shinn, Cassano, Gopinath, Narasimhan, and Yao]{shinn2024reflexion}
Noah Shinn, Federico Cassano, Ashwin Gopinath, Karthik Narasimhan, and Shunyu Yao.
\newblock Reflexion: Language agents with verbal reinforcement learning.
\newblock \emph{Advances in Neural Information Processing Systems}, 36, 2024.

\bibitem[Shridhar et~al.(2020)Shridhar, Yuan, C{\^o}t{\'e}, Bisk, Trischler, and Hausknecht]{shridhar2020alfworld}
Mohit Shridhar, Xingdi Yuan, Marc-Alexandre C{\^o}t{\'e}, Yonatan Bisk, Adam Trischler, and Matthew Hausknecht.
\newblock Alfworld: Aligning text and embodied environments for interactive learning.
\newblock \emph{arXiv preprint arXiv:2010.03768}, 2020.

\bibitem[Song et~al.(2022)Song, Fu, Wang, Sun, Wang, and Zhou]{song2022quantum}
Qingyuan Song, Weiping Fu, Wen Wang, Yuan Sun, Denggui Wang, and Jincao Zhou.
\newblock Quantum decision making in automatic driving.
\newblock \emph{Scientific reports}, 12\penalty0 (1):\penalty0 11042, 2022.

\bibitem[Tan et~al.(2024)Tan, Zhang, Liu, Zheng, Wang, and An]{tantrue}
Weihao Tan, Wentao Zhang, Shanqi Liu, Longtao Zheng, Xinrun Wang, and Bo~An.
\newblock True knowledge comes from practice: Aligning large language models with embodied environments via reinforcement learning.
\newblock In \emph{The Twelfth International Conference on Learning Representations}, 2024.

\bibitem[Team(2024)]{team2024qwen2}
Qwen Team.
\newblock Qwen2 technical report.
\newblock \emph{arXiv preprint arXiv:2407.10671}, 2, 2024.

\bibitem[Wan et~al.(2024)Wan, Feng, Wen, McAleer, Wen, Zhang, and Wang]{wan2024alphazero}
Ziyu Wan, Xidong Feng, Muning Wen, Stephen~Marcus McAleer, Ying Wen, Weinan Zhang, and Jun Wang.
\newblock Alphazero-like tree-search can guide large language model decoding and training.
\newblock In \emph{Forty-first International Conference on Machine Learning}, 2024.

\bibitem[Wang et~al.(2024{\natexlab{a}})Wang, Long, Xiao, Cai, Wu, Meng, Wang, and Zhou]{wang2024biorag}
Chengrui Wang, Qingqing Long, Meng Xiao, Xunxin Cai, Chengjun Wu, Zhen Meng, Xuezhi Wang, and Yuanchun Zhou.
\newblock Biorag: A rag-llm framework for biological question reasoning.
\newblock \emph{arXiv preprint arXiv:2408.01107}, 2024{\natexlab{a}}.

\bibitem[Wang et~al.(2025)Wang, Shi, Hu, Ma, Liu, Wang, Yao, Liu, Ge, and Zhang]{wang2025large}
Jiaqi Wang, Enze Shi, Huawen Hu, Chong Ma, Yiheng Liu, Xuhui Wang, Yincheng Yao, Xuan Liu, Bao Ge, and Shu Zhang.
\newblock Large language models for robotics: Opportunities, challenges, and perspectives.
\newblock \emph{Journal of Automation and Intelligence}, 4\penalty0 (1):\penalty0 52--64, 2025.

\bibitem[Wang et~al.(2024{\natexlab{b}})Wang, Li, Shao, Xu, Dai, Li, Chen, Wu, and Sui]{wang2024math}
Peiyi Wang, Lei Li, Zhihong Shao, Runxin Xu, Damai Dai, Yifei Li, Deli Chen, Yu~Wu, and Zhifang Sui.
\newblock Math-shepherd: Verify and reinforce llms step-by-step without human annotations.
\newblock In \emph{Proceedings of the 62nd Annual Meeting of the Association for Computational Linguistics (Volume 1: Long Papers)}, pp.\  9426--9439, 2024{\natexlab{b}}.

\bibitem[Wang et~al.(2022)Wang, Wei, Schuurmans, Le, Chi, Narang, Chowdhery, and Zhou]{wang2022self}
Xuezhi Wang, Jason Wei, Dale Schuurmans, Quoc Le, Ed~Chi, Sharan Narang, Aakanksha Chowdhery, and Denny Zhou.
\newblock Self-consistency improves chain of thought reasoning in language models.
\newblock \emph{arXiv preprint arXiv:2203.11171}, 2022.

\bibitem[Wiratunga et~al.(2024)Wiratunga, Abeyratne, Jayawardena, Martin, Massie, Nkisi-Orji, Weerasinghe, Liret, and Fleisch]{wiratunga2024cbr}
Nirmalie Wiratunga, Ramitha Abeyratne, Lasal Jayawardena, Kyle Martin, Stewart Massie, Ikechukwu Nkisi-Orji, Ruvan Weerasinghe, Anne Liret, and Bruno Fleisch.
\newblock Cbr-rag: case-based reasoning for retrieval augmented generation in llms for legal question answering.
\newblock In \emph{International Conference on Case-Based Reasoning}, pp.\  445--460. Springer, 2024.

\bibitem[Yan et~al.(2024)Yan, Guo, Lou, Wang, Zhang, and Du]{yan2024efficient}
Xue Yan, Jiaxian Guo, Xingzhou Lou, Jun Wang, Haifeng Zhang, and Yali Du.
\newblock An efficient end-to-end training approach for zero-shot human-ai coordination.
\newblock \emph{Advances in Neural Information Processing Systems}, 36, 2024.

\bibitem[Yan et~al.(2025)Yan, Song, Feng, Yang, Zhang, Ammar, and Wang]{yanefficient}
Xue Yan, Yan Song, Xidong Feng, Mengyue Yang, Haifeng Zhang, Haitham~Bou Ammar, and Jun Wang.
\newblock Efficient reinforcement learning with large language model priors.
\newblock In \emph{The Thirteenth International Conference on Learning Representations}, 2025.

\bibitem[Yao et~al.(2022)Yao, Zhao, Yu, Du, Shafran, Narasimhan, and Cao]{yao2022react}
Shunyu Yao, Jeffrey Zhao, Dian Yu, Nan Du, Izhak Shafran, Karthik Narasimhan, and Yuan Cao.
\newblock React: Synergizing reasoning and acting in language models.
\newblock \emph{arXiv preprint arXiv:2210.03629}, 2022.

\bibitem[Yao et~al.(2023)Yao, Yu, Zhao, Shafran, Griffiths, Cao, and Narasimhan]{yao2023tree}
Shunyu Yao, Dian Yu, Jeffrey Zhao, Izhak Shafran, Thomas~L Griffiths, Yuan Cao, and Karthik Narasimhan.
\newblock Tree of thoughts: Deliberate problem solving with large language models.
\newblock \emph{arXiv preprint arXiv:2305.10601}, 2023.

\bibitem[Yao et~al.(2024)Yao, Yu, Zhao, Shafran, Griffiths, Cao, and Narasimhan]{yao2024tree}
Shunyu Yao, Dian Yu, Jeffrey Zhao, Izhak Shafran, Tom Griffiths, Yuan Cao, and Karthik Narasimhan.
\newblock Tree of thoughts: Deliberate problem solving with large language models.
\newblock \emph{Advances in Neural Information Processing Systems}, 36, 2024.

\bibitem[Yu et~al.(2023)Yu, Gileadi, Fu, Kirmani, Lee, Arenas, Chiang, Erez, Hasenclever, Humplik, et~al.]{yu2023language}
Wenhao Yu, Nimrod Gileadi, Chuyuan Fu, Sean Kirmani, Kuang-Huei Lee, Montserrat~Gonzalez Arenas, Hao-Tien~Lewis Chiang, Tom Erez, Leonard Hasenclever, Jan Humplik, et~al.
\newblock Language to rewards for robotic skill synthesis.
\newblock In \emph{Conference on Robot Learning}, pp.\  374--404. PMLR, 2023.

\bibitem[Zhou et~al.(2025)Zhou, Chen, Guo, Yan, Lee, Wang, Lee, Zhang, Shao, Yang, et~al.]{zhou2025agentfly}
Huichi Zhou, Yihang Chen, Siyuan Guo, Xue Yan, Kin~Hei Lee, Zihan Wang, Ka~Yiu Lee, Guchun Zhang, Kun Shao, Linyi Yang, et~al.
\newblock Agentfly: Fine-tuning llm agents without fine-tuning llms.
\newblock \emph{arXiv preprint arXiv:2508.16153}, 2025.

\bibitem[Zhou et~al.(2024)Zhou, Du, and Li]{zhou2024reflect}
Runlong Zhou, Simon Du, and Beibin Li.
\newblock Reflect-rl: Two-player online rl fine-tuning for lms.
\newblock In \emph{Proceedings of the 62nd Annual Meeting of the Association for Computational Linguistics (Volume 1: Long Papers)}, pp.\  995--1015, 2024.

\end{thebibliography}
